%% file: main_arxiv.tex
\documentclass[runningheads]{llncs}

\usepackage{eccv}

\usepackage{eccvabbrv}

\usepackage{graphicx}
\usepackage{booktabs}

\usepackage[accsupp]{axessibility}  

\usepackage[breaklinks,colorlinks,citecolor=eccvblue]{hyperref}

\usepackage{orcidlink}

\input{preamble}

\begin{document}

\title{\net: Monocular 3D Object Tracking Using Sparse Supervision} 


\author{Nikhil Gosala \inst{1} \and
B Ravi Kiran \inst{2} \and
Senthil Yogamani \inst{3} \and
Abhinav Valada \inst{1}}

\authorrunning{N.~Gosala et al.}

\institute{University of Freiburg, Germany \and
Qualcomm SARL France \and 
Automated Driving, Qualcomm Technologies, Inc., USA}

\maketitle

\input{sections/00_abstract}    
\input{sections/01_introduction}

\input{sections/02_related-work}
\input{sections/03_technical-approach}

\input{sections/04_experiments}

\input{sections/05_conclusion}

\section*{Acknowledgements}
This work was partially funded by Qualcomm Technologies, Inc. and a hardware grant from NVIDIA.

%
%
{
\bibliographystyle{splncs04}
\bibliography{references}
}

\input{sections/06_supplementary}

\end{document}

%% file: preamble.tex
\usepackage{siunitx}
\usepackage{multirow}
\usepackage{pifont}
\usepackage[table]{xcolor}
\usepackage[export]{adjustbox}
\usepackage{float}
\usepackage{microtype}
\usepackage{balance}
\usepackage{flushend}

\newcommand{\net}{Sparse3DTrack}

\newcommand{\cmark}{\text{\ding{51}}}
\newcommand{\xmark}{\text{\ding{55}}}

\newcommand{\graycell}[1]{\cellcolor{gray!40}{#1}}

\newcommand{\tabref}[1]{~\cref{#1}}
\newcommand{\secref}[1]{~\cref{#1}}
\newcommand{\figref}[1]{~\cref{#1}}

%% file: sections/00_abstract.tex
\begin{abstract}
Monocular 3D object tracking aims to estimate temporally consistent 3D object poses across video frames, enabling autonomous agents to reason about scene dynamics. However, existing state-of-the-art approaches are fully supervised and rely on dense 3D annotations over long video sequences, which are expensive to obtain and difficult to scale. 
In this work, we address this fundamental limitation by proposing the first sparsely supervised framework for monocular 3D object tracking. Our approach decomposes the task into two sequential sub-problems: 2D query matching and 3D geometry estimation. Both components leverage the spatio-temporal consistency of image sequences to augment a sparse set of labeled samples and learn rich 2D and 3D representations of the scene.
Leveraging these learned cues, our model automatically generates high-quality 3D pseudolabels across entire videos, effectively transforming sparse supervision into dense 3D track annotations. This enables existing fully-supervised trackers to effectively operate under extreme label sparsity. Extensive experiments on the KITTI and nuScenes datasets demonstrate that our method significantly improves tracking performance, achieving an improvement of up to $15.50$ p.p. while using at most four ground truth annotations per track.
\end{abstract}

%% file: sections/01_introduction.tex
\section{Introduction}
\label{sec:introduction}

Monocular 3D object tracking is a fundamental challenge in computer vision that enables autonomous agents to interact with dynamic environments in real time by estimating the 3D poses and motions of surrounding objects. Accurate 3D tracking is critical for safe navigation and provides essential priors for downstream tasks such as motion forecasting~\cite{schischka2026open,wei2025parkdiffusion} and path planning~\cite{schmidt2025graphpilot,schmalstieg2023learning}. 
Existing monocular 3D object tracking approaches follow a fully supervised training paradigm and thus rely on large amounts of densely annotated 3D bounding boxes and tracks~\cite{cit:track_centertrack, cit:track_mot3d-qd3dt, buchner20223d}. However, acquiring such annotations is both resource-intensive and financially prohibitive, and often requires additional modalities such as LiDAR or Radar to ensure geometric accuracy. Simulation environments and sim-to-real transfer partially alleviate this burden, but their efficacy is limited due to the significant domain gaps that degrade real-world performance. 

 \begin{figure}[t]
    \centering
    {\includegraphics[width=0.8\linewidth]{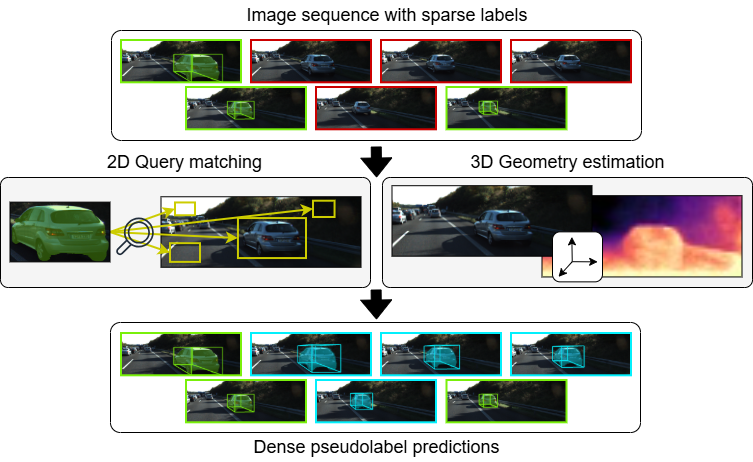}}
    \caption{\net: The first sparsely supervised approach for monocular 3D object tracking. Our framework couples spatio-temporal consistency of image sequences with sparse annotations (green) to generate high-quality 3D tracking pseudolabels (blue).}
    \label{fig:teaser}
    \vspace{-10pt}
\end{figure}

To address the aforementioned limitations, we propose~\textit{\net}, the first framework for sparsely supervised monocular 3D object tracking. The core idea of our approach is to augment a small number of human annotations with rich spatio-temporal context from image sequences to generate accurate 3D pseudolabels, thereby transforming sparse supervision into dense annotations. Our framework thus establishes a new paradigm for monocular 3D object tracking, enabling existing trackers to be trained under extreme label sparsity. 
 
\net~decomposes the overall problem into two sequential sub-tasks, \textit{2D query matching} and \textit{3D geometry estimation}, as shown in~\figref{fig:teaser}. The 2D query matching module establishes object correspondences across frames by measuring feature similarity between a query region and target images. To mitigate label sparsity, it incorporates a data mining mechanism that exploits the spatio-temporal consistency of unlabeled neighboring frames to automatically generate additional training samples. This enables robust object localization despite significant variations in appearance, illumination, and viewpoint. Building on these localizations, the 3D geometry estimation module regresses the object's 3D position by fusing these matched features with depth priors from a self-supervised depth network. It also addresses the ill-posed 3D yaw estimation problem by decomposing orientation prediction into keypoint regression and depth reasoning, thus enabling accurate 3D pose recovery even under extreme label sparsity. By learning to independently reason about each sub-task and subsequently fusing their outputs, our framework produces accurate and temporally consistent 3D track pseudolabels from only a handful of human annotations.
Extensive experiments on the KITTI and nuScenes datasets demonstrate that~\net~generates high-quality pseudolabels and significantly improves the performance of existing monocular 3D trackers, achieving an improvement of up to $15.50$ p.p. while using at most four annotations per track. Moreover, a comprehensive ablation study further validates the effectiveness of each proposed component, highlighting the efficacy of our framework. 

Our main contributions can be summarized as follows:
\begin{enumerate}
    \item The first sparsely supervised framework for monocular 3D object tracking.
    \item A 2D query matching module with a novel sample mining mechanism to localize a query region in target images using only sparse annotations.
    \item A 3D geometry estimation module with a novel yaw formulation to recover the 3D pose of localized queries under extremely sparse supervision.
    \item A false negative compensation module to handle missed detections in the generated 3D pseudolabels.
    \item Four competitive baselines for the novel task of sparsely supervised 3D monocular object tracking.
\end{enumerate}

%% file: sections/02_related-work.tex
\section{Related Work}
\label{sec:related-work}

\noindent\textbf{Monocular 2D Object Tracking} is a fundamental challenge in computer vision that associates and localizes objects across an image sequence in the pixel space. The following section summarizes prior work using fully supervised and label-efficient training regimes.
{\parskip=2pt
\noindent\textit{Fully supervised tracking} follows two alternate paradigms: tracking-by-detection (TBD) and joint detection and tracking (JDT). TBD frameworks associate detections across frames using cosine similarity~\cite{cit:track_deepsort}, Gaussian mixture-based motion modeling alongside appearance similarity~\cite{cit:track_hta}, and global associations using Gaussian-smoothed tracks~\cite{cit:track_strongsort}. LSTMs have also been used to enforce temporal consistency and capture long-term dependencies~\cite{cit:track_bilinear-lstm, cit:track_lstm, cit:track_lstm-pooling}. The JDT paradigm introduced by~\cite{cit:track_tracktor} performs detection and tracking in a unified step, with subsequent work incorporating optical flow to aid track propagation~\cite{cit:track_tracktor-flow}. ~\cite{cit:track_centertrack} introduced proposal-free tracking using object center heatmaps to reduce computational complexity and boost tracking speed. Recent transformer-based models enable end-to-end spatio-temporal reasoning and are the current state-of-the-art~\cite{cit:track_transtrack, cit:track_trackformer, cit:track_transmot, cit:track_transcenter}.
}

{\parskip=2pt
\noindent\textit{Label-efficient tracking} reduces the dependency on dense manual annotations by synthetically generating appearance variations using cut-transform-paste~\cite{cit:track_li2023self} or by auto-labeling image sequences using spatio-temporal constraints~\cite{cit:track_video-labeling}. \cite{cit:track_spot-sparse-supervision} advances label-efficient tracking by leveraging a teacher-student framework with spatio-temporal constraints, while~\cite{lang2023self} trains associations across multiple frames to improve robustness and consistency for object re-identification. However, these approaches primarily target scenes with limited depth variation and low object clutter. In contrast, our 2D query matching module presents a label-efficient formulation for autonomous driving, enabling tracking under severe occlusions, clutter, and large depth variations by localizing objects using both 2D bounding boxes and object masks.}

\noindent\textbf{Monocular 3D Object Tracking} associates and localizes objects in the 3D space using only images from a single camera. The following section outlines prior work under fully supervised and label-efficient settings.

{\parskip=2pt
\noindent\textit{Fully supervised} monocular 3D object tracking is primarily formulated under the TBD paradigm, where current frame 3D detections are associated with existing trajectories. Early works explored distance-based association metrics such as IoU~\cite{cit:track_mot3d-iou}, L2~\cite{cit:track_mot3d-l2} and GIoU~\cite{cit:track_mot3d-giou}. Later approaches, in contrast, incorporate learnable motion dynamics and appearance cues to enhance robustness under occlusion and motion ambiguity~\cite{cit:track_mot3d-qd3dt, cit:track_mot3d-fantrack, cit:track_mot3d-deft}. Recent transformer-based frameworks extend this paradigm by leveraging attention to learn object similarity~\cite{cit:track_mot3d-time3d}. Further, recent approaches encode appearance, pose, and motion information into a single track query, and thus enable end-to-end optimization across detection and tracking~\cite{cit:track_mot3d-motr, cit:track_mot3d-mutr3d, cit:track_mot3d-adatrack,kappeler2025bridging}. 
}

{\parskip=2pt
\noindent\textit{Label-efficient approaches} in 3D tracking have been limited to the LiDAR domain with~\cite{cit:track_lidar-auto4d} proposing an iterative mechanism to smooth noisy 3D detections, and \cite{cit:track_lidar-labelformer} employing self-attention to refine object trajectories and infer accurate poses. Recently, \cite{cit:track_lidar-sparse-autolabels} explored generating 3D track pseudolabels from point clouds using sparse supervision. 
However, no approach has yet addressed label-efficient monocular 3D object tracking, where insufficient depth annotations hinder 3D pose recovery. \net~is thus the first framework to tackle this challenge, thereby enabling monocular 3D object tracking using extremely sparse supervision.
}

%% file: sections/03_technical-approach.tex
\section{Technical Approach}
\label{sec:technical-approach}

In this section, we present our novel \net~framework for sparsely supervised monocular 3D object tracking. The core idea of our approach is to divide the problem into two sequential sub-tasks: \textit{2D query matching} and \textit{3D geometry estimation}. The 2D query matching module augments sparse annotations with spatio-temporal consistency of image sequences to learn object correspondence across unlabeled images using an appearance-based loss ($\mathcal{L}_\text{match}$, \secref{subsec:2d-query-matching}). Subsequently, the 3D geometry estimation module learns the 3D pose of the localized object by coupling the appearance features with self-supervised depth information using a geometry regression loss ($\mathcal{L}_\text{geom}$, \secref{subsec:3d-geometry-estimation}). This two-stage protocol enables \net~to generate dense 3D pseudolabels from sparse annotations, thereby enabling the training of state-of-the-art monocular 3D object trackers. The total loss of the network is thus computed as:
\begin{align}
    \mathcal{L} = \mathcal{L}_\text{match} + \mathcal{L}_\text{geom}.
\end{align}

In the following sections, we describe the overall network architecture and provide insights into our 2D query matching and 3D geometry estimation modules. Here, an image with a query annotation is denoted \textit{source image}, the region of interest within the query bounding box is termed \textit{template image}, and the unlabeled image in which we want to find the corresponding object is termed \textit{target image}. Further, the annotated object in the source image is referred to as the \textit{query object} while the associated prediction in the target image is termed \textit{target object}. In all subsequent notations, the subscript $k$ refers to an instance at timestep $t_k$.

\begin{figure*}
    \centering
    {\includegraphics[width=\linewidth]{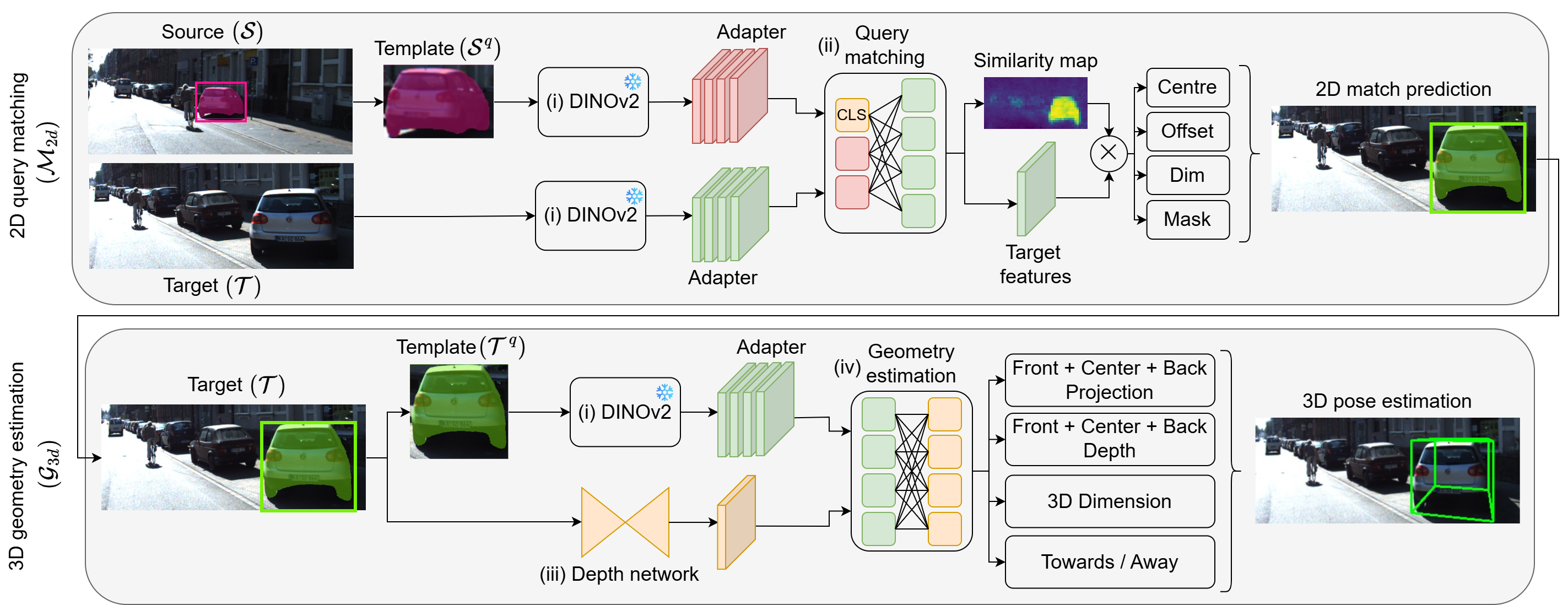}}
    \caption{Overview of our \net~framework for sparsely supervised monocular 3D object tracking. \net~decomposes the overall problem into two sequential sub-tasks, namely, \textit{2D query matching} which finds object correspondences across frames and \textit{3D geometry estimation} which estimates the 3D pose and dimensions of the localized object. Our framework is extremely label-efficient and enables object tracking using at most four annotations per track.}
    \label{fig:network-architecture}
\end{figure*}

\subsection{Network Architecture}
\label{subsec:network-architecture}

Our network, illustrated in \figref{fig:network-architecture}, encapsulates four key modules: (i) a DINOv2~\cite{cit:dinov2} image encoder to extract features from an input image, (ii) a transformer-based 2D query matching module to localize the query object in the target image using feature similarity between the template and target images, (iii) a convolution-based self-supervised depth estimation network to generate intermediate depth features, and (iv) a transformer-based 3D geometry estimation module to fuse the target object and depth features, and subsequently regress the 3D object pose and size estimates. During inference, the 2D query matching module first localizes the target object using bounding box and mask predictions. The 3D geometry estimation network then uses these outputs to estimate the 3D pose of the object.

\subsection{2D Query Matching}
\label{subsec:2d-query-matching}
The goal of the 2D query matching module $\mathcal{M}_{2d}$ is to establish object correspondence between a pair of images. Given a query object from a source image, this module localizes the corresponding object in the target image, thereby propagating object correspondence across an image sequence and facilitating 2D object tracking.

\begin{figure}[t]
\begin{minipage}{0.55\linewidth}
\centering
    {\includegraphics[width=\linewidth]{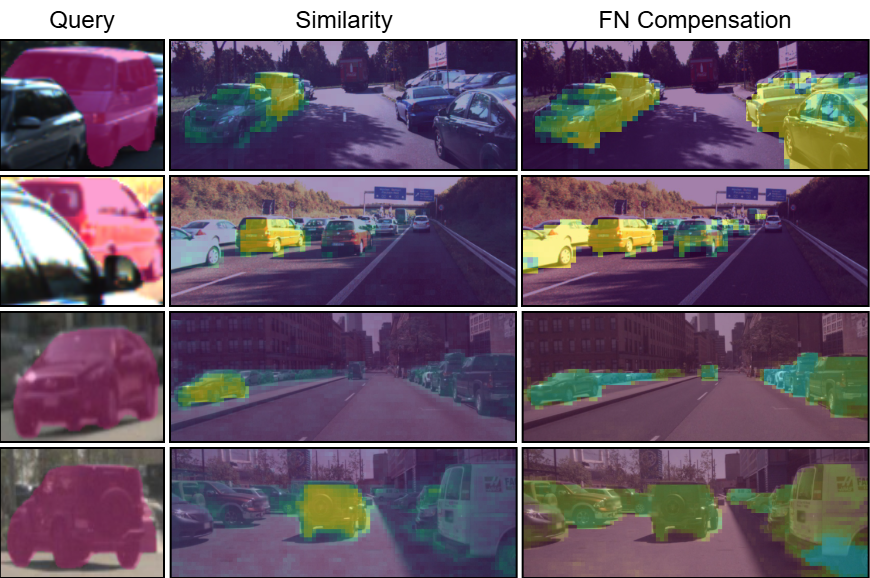}}
    \caption{Similarity and FNComp heatmaps computed by the $\mathcal{M}_{2d}$ and FNComp modules, respectively. Note that the similarity map focuses only on the query object in the target image, while the FNComp heatmap highlights all vehicles in the scene.}
    \label{fig:app2d-similarity-outputs}
\end{minipage}
\hfill
\begin{minipage}{0.4\linewidth}
\centering
    \includegraphics[width=\linewidth]{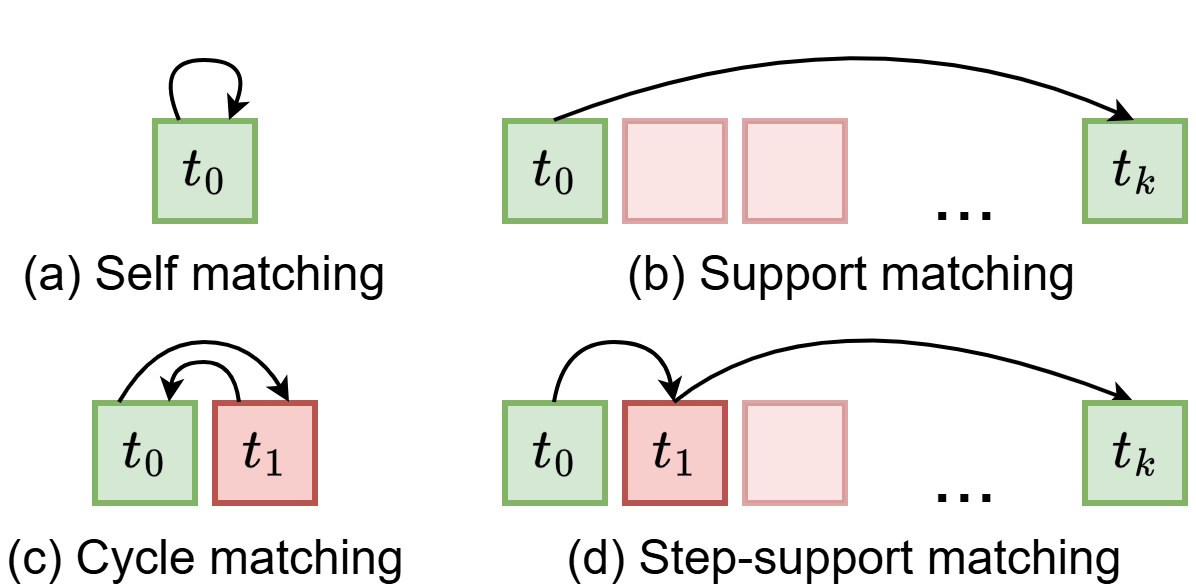}
    \caption{Illustration of data mining strategies used to tackle label sparsity in $\mathcal{M}_{2d}$. In this figure, green and red boxes represent labeled and unlabeled image frames, respectively.}
    \label{fig:app2d-training-strategies}
\end{minipage}
\end{figure}

We first crop and mask the source image $\mathcal{S}$ using the query box $q^{b}$ and query object mask $q^{m}$ to extract the template image $\mathcal{S}^q$. We then process $\mathcal{S}^q$ using an image encoder and an adapter network to generate the query object features. In a parallel branch, we pass the target image $\mathcal{T}$ through the same image encoder and adapter networks to generate the target image features. We then compute the similarity between the two feature maps using a similarity estimation module comprising a self-attention layer for the query object features and a cross-attention layer between the two feature maps. 
This module outputs a similarity map that determines regions in the target image similar in appearance to the query object, i.e., regions most likely to contain the query object, as shown in \figref{fig:app2d-similarity-outputs}. We supervise the similarity estimation module with a contrastive infoNCE loss, where positive samples are obtained from the ground truth target object annotations and hard negative samples are mined from highly confident yet incorrect similarity predictions.
The infoNCE loss is computed as
\begin{equation}
    \mathcal{L}^\text{sim}_\text{match} = \text{CE}(sim(f_{S^q}, f_{T}), y),
\end{equation}
where $f_{S^q}$ and $f_{T}$ denote the sampled features from template and target respectively, $y$ represents the ground truth vector determining whether an entry belongs to the target object, $sim(\cdot, \cdot)$ refers to the cosine similarity function, and $\text{CE}(\cdot, \cdot)$ denotes the cross entropy loss between two tensors. 

We subsequently weight the target features using the similarity map to generate a similarity-aware feature representation, then process it with independent convolutional heads to regress the center heatmap, center offset, and bounding box dimensions.
We also use the mask head outlined in~\cite{cit:track_centermask} to compute the target object mask. We supervise the center heatmap using focal loss ($\mathcal{L}^\text{cen}_\text{match}$), center offset and box dimensions using $L1$ loss ($\mathcal{L}^\text{off}_\text{match}, \mathcal{L}^\text{dim}_\text{match}$), and object mask using BCE loss ($\mathcal{L}^\text{mask}_\text{match}$).
The final loss of the 2D query matching module is thus computed as
\begin{equation}
    \mathcal{L}_\text{match} = \mathcal{L}^\text{sim}_\text{match} + \mathcal{L}^\text{cen}_\text{match} + \mathcal{L}^\text{dim}_\text{match} + \mathcal{L}^\text{off}_\text{match} + \mathcal{L}^\text{mask}_\text{match}.
\end{equation}

{\parskip=2pt
\noindent\textbf{\textit{Tackling Label Sparsity:}} We tackle extreme label sparsity by designing a data augmentation pipeline to generate rich training samples from limited labels. Specifically, we mine unique source and target pairs using four strategies: (a) self matching, (b) support matching, (c) cycle matching, and (d) step-support matching as shown in \figref{fig:app2d-training-strategies}. 
The self-matching strategy uses a single labeled image $I_0$ as both the source and target. Matching a query object to itself provides an optimal training signal to the similarity estimation module, enabling the network to learn perfect feature correspondences. In contrast, support matching uses two distinct labeled images, $I_0$ and $I_k$ as source and target, respectively. This strategy trains the similarity module to generalize object matching across varying views and appearances. Together, these supervised strategies constrain the network parameters, preventing them from learning incorrect relationships. 

Cycle matching and step-support matching leverage unlabeled image neighbors to augment network training. These approaches use an unlabeled image as an intermediate waypoint whose pseudolabel prediction is subsequently used to establish an association with a labeled image. To this end, both strategies begin by using a labeled image $I_0$ as the source to localize the query object in an unlabeled neighbor, which then becomes the new source. For cycle matching, we match the neighbor back to the original source image $I_0$, teaching the network to maintain appearance consistency across multiple prediction steps. In contrast, step-support matching matches the neighbor to a different labeled image $I_k$, enhancing the network's ability to maintain long-range correspondence. By using unlabeled intermediate waypoints, we increase the number of training samples and make the network more robust to prediction noise. This skill is key for real-world inference, where tracks are propagated based on prior predictions.}

\subsection{3D Geometry Estimation}
\label{subsec:3d-geometry-estimation}
The 3D geometry estimation module $\mathcal{G}_{3d}$ estimates the 3D pose of the target object using a monocular RGB image. This module lifts the 2D object correspondence obtained from $\mathcal{M}_{2d}$ into 3D space, thereby enabling 3D object tracking with sparse supervision. However, monocular 3D estimation is fundamentally ill-posed, whose challenge is further exacerbated by extreme label sparsity, which prevents depth reasoning via supervised learning. This module circumvents this limitation by leveraging strong depth priors from a self-supervised depth network alongside sparse annotations to promote 3D understanding.

We first crop and mask the query object from the target image $\mathcal{T}$ using predictions from $\mathcal{M}_{2d}$, then process it with an image encoder and an adapter module to generate the query object features.
In parallel, we compute the depth features of the query object by processing $\mathcal{T}$ with a self-supervised depth network and then cropping the resulting depth features with the query bounding box. We then fuse these features using an encoder-decoder transformer comprising self- and cross-attention layers between the image and depth features to generate a feature vector for the target object. 
We then regress the 3D center and dimensions of the object using two separate heads and supervise them using L1 loss. 

{\parskip=2pt
\noindent\textbf{\textit{3D Yaw Estimation:}} Monocular 3D yaw estimation is also an ill-posed problem due to the ambiguity in inferring orientation from 2D projection. Traditional approaches relying on either direct regression~\cite{cit:detection_yaw-regression} or bin classification and offset estimation~\cite{cit:track_mot3d-qd3dt, cit:track_centertrack} require large-scale supervision and are thus unsuitable in our label-efficient setting. We overcome this limitation by reformulating 3D yaw estimation as a function of 3D keypoint regression on the target object. To this end, our network learns to predict the 2D projections and corresponding depth values of the object's front, center, and back points. We also predict a binary direction variable $d \in \{\text{towards}, \text{away}\}$ which determines whether the object faces towards or away from the ego vehicle, thus resolving the direction ambiguity.
The 3D yaw is then computed as
\begin{equation}
    \theta = 
    \begin{cases}
        -\arctan(z_f - z_c, x_f - x_c), & \text{if } d = \text{towards} \\
        \pi - \arctan(z_b - z_c, x_b - x_c) & \text{if } d = \text{away},
    \end{cases}
\end{equation}
where $(x_f, y_f, z_f)$, $(x_c, y_c, z_c)$, and $(x_b, y_b, z_b)$ represent the 3D coordinates of the object's front, center, and back points computed by lifting the corresponding 2D projections using their depth values. This formulation decomposes 3D yaw estimation into 2D projection regression and depth estimation, where the former is well-defined and the latter can be estimated alongside 3D center estimation using self-supervised depth features. We train the 3D coordinates of the front and back points using the L1 loss function, and the binary direction variable using the BCE loss function.  
The final geometry loss $\mathcal{L}_\text{geom}$ is thus computed as
\begin{equation}
    \mathcal{L}_\text{geom} = \mathcal{L}^\text{keypoint}_\text{geom} + 
    \mathcal{L}^\text{depth}_\text{geom} + \mathcal{L}^\text{dim}_\text{geom} + \mathcal{L}^\text{dir}_\text{geom},
\end{equation}
where $\mathcal{L}^\text{keypoint}_\text{geom}$ represents the regression loss for the center, front, and back keypoints of object, $\mathcal{L}^\text{depth}_\text{geom}$ denotes the L1 loss for the keypoint depths, $\mathcal{L}^\text{dim}_\text{geom}$ refers to the loss for the object dimensions, and $\mathcal{L}^\text{dir}_\text{geom}$ denotes the classification loss for the binary direction variable.}

\subsection{Pseudolabel Generation}
\label{subsec:plabel-generation}
We generate the dense 3D tracking pseudolabels following a two-stage protocol. We first execute $\mathcal{M}_{2d}$ to generate a 2D bounding box and mask prediction for the target object, and then process these outputs using $\mathcal{G}_{3d}$ to predict the object's 3D position and orientation.~\net~initializes an object track using a sparse label and iteratively propagates the track across the entire image sequence. We mitigate prediction errors using heatmap confidences in $\mathcal{M}_{2d}$ wherein we discard predictions having confidence less than $0.5$, and update the source image only when the match confidence is higher than $0.75$. 
We generate pseudolabels in both forward and backward directions and merge the outputs based on their relative confidence to produce the final set of high-quality 3D pseudolabels.

{\parskip=2pt
\noindent\textbf{\textit{Handling False Negatives:}} The pseudolabels generated by~\net~may suffer from false negatives (FNs) where valid objects in the scene lack corresponding labels. Downstream tracking models erroneously treat these FN regions as true negatives, in turn degrading the tracking model. To address this problem, we propose a False Negative Compensation~(FNComp) module that identifies image regions that may contain unlabeled objects and excludes them from downstream model training. The FNComp architecture mirrors that of $\mathcal{M}_{2d}$, and predicts 2D heatmaps indicating the probability of spatial regions containing valid objects (see \figref{fig:app2d-similarity-outputs}). In contrast to $\mathcal{M}_{2d}$, negative samples are chosen from low confidence regions, enabling the FNComp module to effectively disambiguate between foreground and background. The resulting heatmap is then used to downweight the negative losses in regions having high probabilities of FNs, thereby avoiding penalizing the network for correctly predicting unlabeled objects.
}

%% file: sections/04_experiments.tex
\section{Experimental Results}
\label{sec:experiments}
In this section, we evaluate our novel pseudolabel generation framework through both quantitative and qualitative analyses and conduct extensive ablation experiments to demonstrate the efficacy of our proposed contributions.

\subsection{Datasets}
We evaluate~\net~on the vehicle class from two automotive tracking datasets, namely, KITTI Tracking~\cite{cit:dataset-kitti} and nuScenes~\cite{cit:dataset-nuscenes}. We generate the sparse labels by following the sampling strategy described in ~\cite{cit:track_spot-sparse-supervision} and sampling at most $4$ labels per track, thereby reducing the overall annotation density in KITTI and nuScenes by $92.75\%$ and $74.34\%$, respectively. To reflect real-world human annotation, we only sample annotations where objects are not heavily occluded, i.e., annotations with occlusion levels of (0, 1) in KITTI and visibility values of (2, 3, 4) in nuScenes. We adopt the train-val splits outlined in~\cite{cit:track_mot3d-qd3dt}, resulting in $13$ training and $8$ validation sequences for KITTI, and $700$ training and $150$ validation sequences for nuScenes.

\subsection{Training Protocol}
We train~\net~on images of size $448\times1344$ pixels on both datasets. We augment training images using random horizontal flips and random perturbations of brightness, contrast, saturation, and hue. We train both the query matching and geometry estimation networks using the sparse labels on the KITTI dataset for a total of $500$ epochs using an initial learning rate (LR) of $0.001$, which is scaled by $0.5$ at epochs $350$, $400$, and $450$. On nuScenes, we train our model for $100$ epochs using the same LR, which is halved at epochs $70$, $80$, and $90$. We optimize our model using SGD with a batch size of $12$, momentum of $0.9$, and weight decay of $1e^{-4}$.

The self-supervised depth network follows the implicit fields formulation described in~\cite{cit:bev_letsmap} and is trained independently for $30$ epochs using an LR of $0.005$, which is reduced by a factor of $10$ at epochs $16$ and $22$. We optimize the depth network using SGD with a batch size of $12$. Owing to noisy IMU labels in the KITTI tracking split, we train the depth network on the KITTI odometry dataset, which contains SLAM-optimized poses, and deploy the resulting model on the tracking split. In contrast, nuScenes provides accurate ego poses and does not require this adjustment.

\begin{table*}
\centering
\tiny
\caption{Evaluation of our approach on the KITTI Tracking and nuScenes datasets. The results of our model are highlighted in grey, and the best values are indicated in bold. Here, ``Sparse'' and ``Oracle'' refer to the cases when the base network is trained using only the sparse labels (lower bound) and $100\%$ ground truth labels (upper bound), respectively. All values are represented in [$\%$].}
\label{tab:quant-pseudolabel-eval}
\begin{tabular}{c|c|c|ccc|cccc}
 \toprule
& & & \multicolumn{3}{c|}{\textbf{KITTI Tracking}} & \multicolumn{4}{c}{\textbf{nuScenes}} \\
\cmidrule{4-10}
 \textbf{Base} & \textbf{2D} & \textbf{3D} & \textbf{MOTA}~$\uparrow$ & \textbf{MOTP}~$\downarrow$ & \textbf{IDF1}~$\uparrow$ & 
 \textbf{AMOTA}~$\uparrow$ & \textbf{AMOTP}~$\downarrow$ & \textbf{MOTA}~$\uparrow$ & \textbf{MOTP}~$\downarrow$ \\
 \midrule
 \multirow{10}{*}{\rotatebox[origin=c]{90}{CenterTrack~\cite{cit:track_centertrack}}} & \multicolumn{2}{c|}{Sparse} & 14.21 & 53.15 & 19.97 & 14.06 & 160.24 & 12.20 & 68.82 \\
\cmidrule{2-10}
  & SPOT~\cite{cit:track_spot-sparse-supervision} & FCOS3D~\cite{cit:det3d_fcos3d} & 19.05 & 46.78 & 31.00 & 10.26 & 161.31 & 7.82 & 89.51 \\
 & SPOT~\cite{cit:track_spot-sparse-supervision} & CenterNet~\cite{cit:det3d_centernet} & 21.83 & \textbf{42.74} & 31.33 & 18.33 & \textbf{142.83} & 12.84 & \textbf{62.81} \\
 & \graycell{Ours} & \graycell{Ours} & \graycell{\textbf{29.71}} & \graycell{47.21} & \graycell{\textbf{35.66}} & \graycell{\textbf{23.97}} & \graycell{151.63} & \graycell{\textbf{19.48}} & \graycell{74.14} \\
 \cmidrule{2-10}
 & SAM2~\cite{cit:track_mot2d-sam2} & FCOS3D~\cite{cit:det3d_fcos3d} & 26.31 & 48.77 & 37.80 & 0.71 & 173.24 & 1.31 & 107.13 \\
 & SAM2~\cite{cit:track_mot2d-sam2} & CenterNet~\cite{cit:det3d_centernet} & 27.42 & \textbf{47.28} & 42.20 & 20.00 & 151.70 & 13.91 & 82.28 \\
 & \graycell{SAM2~\cite{cit:track_mot2d-sam2}} & \graycell{Ours} & \graycell{\textbf{35.63}} & \graycell{49.22} & \graycell{\textbf{45.00}} &\graycell{\textbf{29.15}} & \graycell{\textbf{144.00}} & \graycell{\textbf{20.19}} & \graycell{\textbf{81.09}} \\
 \cmidrule{2-10}
 & \multicolumn{2}{c|}{Oracle} & 44.32 & 54.65 & 50.08 & 38.51 & 126.06 & 29.40 & 64.43 \\

 \midrule

 \multirow{10}{*}{\rotatebox[origin=c]{90}{DEFT~\cite{cit:track_mot3d-deft}}} & \multicolumn{2}{c|}{Sparse} & 13.60 & 42.00 & 29.22 & 20.42 & 153.96 & 22.53 & 76.27 \\
\cmidrule{2-10}
 & SPOT~\cite{cit:track_spot-sparse-supervision} & FCOS3D~\cite{cit:det3d_fcos3d} & 20.10 & 43.08 & 42.19 & 21.70 & 154.68 & 17.56 & 88.03 \\
 & SPOT~\cite{cit:track_spot-sparse-supervision} & CenterNet~\cite{cit:det3d_centernet} & 22.81 & \textbf{40.06} & 44.89 & 25.43 & 149.64 & 19.75 & 77.43 \\
 & \graycell{Ours} & \graycell{Ours} & \graycell{\textbf{28.33}} & \graycell{41.65} & \graycell{\textbf{47.63}} & \graycell{\textbf{26.38}} & \graycell{\textbf{148.93}} & \graycell{\textbf{21.26}} & \graycell{\textbf{76.69}} \\
 \cmidrule{2-10}
 & SAM2~\cite{cit:track_mot2d-sam2} & FCOS3D & 23.94 & \textbf{41.38} & 49.19 & 7.14 & 168.09 & 8.00 & 93.42 \\
 & SAM2~\cite{cit:track_mot2d-sam2} & CenterNet~\cite{cit:det3d_centernet} & 24.57 & 43.42 & 53.04 & 27.46 & 149.54 & 22.45 & 85.27 \\
 & \graycell{SAM2~\cite{cit:track_mot2d-sam2}} & \graycell{Ours} & \graycell{\textbf{32.70}} & \graycell{42.97} & \graycell{\textbf{56.78}} & \graycell{\textbf{36.03}} & \graycell{\textbf{141.88}} & \graycell{\textbf{29.59}} & \graycell{\textbf{81.77}} \\
 \cmidrule{2-10}
 & \multicolumn{2}{c|}{Oracle} & 44.02 & 46.75 & 67.32 & 47.37 & 119.51 & 37.15 & 68.72\\

\bottomrule
\end{tabular}
\vspace{-0.1cm}
\end{table*}

\subsection{Quantitative Results}
As we are the first to address monocular 3D object tracking using sparse supervision, we establish a new benchmark for this task. We construct four competitive baselines by combining existing 2D object trackers (SPOT~\cite{cit:track_spot-sparse-supervision} and SAM2~\cite{cit:track_mot2d-sam2}) with 3D detection frameworks (FCOS3D~\cite{cit:det3d_fcos3d} and CenterNet~\cite{cit:det3d_centernet}). We then evaluate the quality of the resulting 3D pseudolabels by comparing the tracking performance using two camera-based 3D object tracking models, namely, CenterTrack~\cite{cit:track_centertrack} and DEFT~\cite{cit:track_mot3d-deft}. Note that SPOT is the latest 2D object tracking framework leveraging sparse supervision, while SAM2 is a state-of-the-art 2D tracking foundation model trained using large corpora of annotated data. These baselines thus allow evaluation of our approach across different label regimes. 

To contextualize these tracking results and quantify the benefit of pseudolabel generation, we additionally report results obtained when these base networks are trained using only sparse annotations (lower bound, ``Sparse'' in \tabref{tab:quant-pseudolabel-eval}) and $100\%$ ground truth labels (upper bound, ``Oracle'' in \tabref{tab:quant-pseudolabel-eval}). To ensure fair comparison, all models are trained using their public codebases. We quantify tracking performance using the MOTA, MOTP, and IDF1 metrics on the KITTI dataset, while we report AMOTA, AMOTP, MOTA, and MOTP metrics for the nuScenes dataset. \tabref{tab:quant-pseudolabel-eval} summarizes the results of these experiments on both the KITTI and nuScenes datasets. For brevity, we describe the results in the following text using only CenterTrack as the base model. Similar conclusions can be drawn when using DEFT as the base tracking model, and this is further elaborated in \secref{sec:supp-quantitate-results} of the supplementary material. 

On the KITTI dataset, we observe that our model significantly outperforms the sparse baseline by more than $100\%$, improving MOTA by $15.50$ p.p., highlighting the benefit of our~\net~framework. We observe that our model also exceeds both SPOT baselines by $7.88$ p.p. on MOTA and $4.33$ p.p. on IDF1, indicating superior localization accuracy and track consistency. SPOT, designed primarily for scenes with static cameras and constant depth planes, fails to generalize to driving sequences involving camera motion and multiple depth planes, resulting in false positives. Moreover, its overconfident tracking behavior maintains object tracks even after they exit the frame, in turn degrading model performance. In contrast, our approach uses the similarity estimation module to handle camera motion and depth variation while terminating out-of-frame tracks, yielding stable 3D tracking pseudolabels across diverse driving conditions. Further, we note that our model surpasses both SAM2 baselines by over $2.29$ p.p., highlighting the benefit of our 3D geometry estimation module. This is further supported by the fact that SAM2 combined with our 3D geometry estimation module achieves the best tracking performance, exceeding all baselines by at least $8.21$ p.p. This improvement stems from the use of a self-supervised depth estimation network to enhance depth reasoning in the face of label sparsity as well as our novel yaw formulation that enables accurate orientation estimation with minimal supervision.

We observe a similar trend on the more challenging nuScenes dataset, where our approach surpasses the sparse baseline by $9.91$ p.p. and outperforms both SPOT variants by $13.71$ p.p. and $5.64$ p.p., demonstrating strong cross-dataset generalization. Our framework also significantly exceeds both the SAM2 baselines despite relying solely on sparse annotations instead of fully supervised foundation model training. Further, combining SAM2 with our label-efficient 3D geometry estimation module yields the best overall result, exceeding all approaches by more than $5$ p.p. on the AMOTA score. 
Together, these results highlight the high quality of 3D pseudolabels generated by~\net, allowing existing monocular 3D object trackers to be trained using extremely limited supervision. 

\subsection{Ablation Study}
In this section, we analyze the impact of various components of our framework by performing an ablation study on the KITTI dataset. Specifically, we perform four ablation experiments to study the impact of (i) our sample mining strategy on 2D query matching, (ii) using a self-supervised depth network and our novel yaw estimation strategy on 3D geometry estimation, (iii) using confidence weighting and applying FN compensation heatmaps on the base network, and (iv) increasing the number of sparse annotations on the overall model performance. For each ablation experiment, we regenerate the 3D tracking pseudolabels with the modified network, train CenterTrack~\cite{cit:track_centertrack} on the new pseudolabels, and report the resulting tracking performance on the MOTA and IDF1 metrics.

{\parskip=2pt
\noindent\textbf{\textit{Sample Mining in $\mathcal{M}_{2d}$}}:
We analyze the impact of each sample mining strategy outlined in \secref{subsec:2d-query-matching} on $\mathcal{M}_{2d}$ by progressively incorporating them into the overall model, as summarized in \tabref{tab:quant-ablation-app2d}. Model A1, trained using only the na\"ive support-matching mechanism, serves as the baseline. Adding the self-matching module in model A2 improves the MOTA and IDF1 scores by $2.85$ p.p. and $1.59$ p.p., respectively, which can be attributed to the enhanced feature grounding offered by self-consistency, which enables optimal feature correspondence learning. Incorporating cycle- and step-support-matching in models A3 and A4 yields further gains of $7.82$ p.p. in MOTA and $8.03$ p.p in IDF1. This demonstrates that using unlabeled image neighbors as intermediate waypoints provides a strong supervision signal and enables rich feature learning in the face of label sparsity.}

{\parskip=2pt
\noindent\textbf{\textit{Self-Supervised Depth and Yaw Estimation in $\mathcal{G}_{3d}$}}: \tabref{tab:quant-ablation-geom3d} analyzes the impact of leveraging self-supervised depth features in $\mathcal{G}_{3d}$. We note that the model using depth features~(D2) outperforms the variant without depth~(D1) by $2.92$ p.p. on the MOTA metric, demonstrating the benefit of depth features for sparse model training. We also observe that IDF1 remains largely unchanged, as the quality of object identification is primarily governed by $\mathcal{M}_{2d}$ and is thus less influenced by depth features.
Further, we quantify the impact of using three different yaw estimation strategies, i.e., regression-based, bin-based, and our proposed approach on the overall model performance. Our method~(Y3) achieves the best performance, exceeding the regression-based~(Y1) and bin-based~(Y2) approaches by $4.59$ p.p. and $3.12$ p.p. in MOTA, respectively. By decomposing the complex 3D yaw estimation task into simpler 2D regression and 3D depth sub-tasks, our approach yields more accurate orientation estimates under sparse supervision. Further, the change in IDF1 is larger than in the depth ablation, since incorrect yaw prediction affects future track propagation via erroneous 2D matching, thus degrading pseudolabel quality.
}

\begin{table}[t]
\tiny
\centering
\setlength\tabcolsep{3pt}
\begin{minipage}[t]{.48\linewidth}
\centering
\caption{Impact of different sample mining strategies on $\mathcal{M}_{2d}$.}
\label{tab:quant-ablation-app2d}
 \begin{tabular}{c|cccc|cc}
 \toprule
 & \textbf{Sup} & \textbf{Self} & \textbf{Cyc} & \textbf{SS} & \textbf{MOTA} & \textbf{IDF1} \\
\midrule
A1 & \cmark & \xmark & \xmark & \xmark & 17.41 & 21.98 \\
A2 & \cmark & \cmark & \xmark & \xmark & 20.26 & 23.57 \\
A3 & \cmark & \cmark & \cmark & \xmark & 22.74 & 27.40 \\
A4 & \cmark & \cmark & \cmark & \cmark & \textbf{28.08} & \textbf{31.60} \\
\bottomrule
\end{tabular}
\end{minipage}%
\hfill
\begin{minipage}[t]{.48\linewidth}
\centering
\caption{Impact of depth network and yaw estimation on $\mathcal{G}_{3d}$.}
\label{tab:quant-ablation-geom3d}
 \begin{tabular}{c|cc|cc}
 \toprule
 & \textbf{Depth} & \textbf{Yaw} & \textbf{MOTA} & \textbf{IDF1} \\
\midrule
D1 & \xmark & Ours & 25.16 & 31.12 \\
D2 & \cmark & Ours & \textbf{28.08} & \textbf{31.60} \\
\midrule
Y1 & \cmark & Regr & 23.49 & 30.08 \\
Y2 & \cmark & Bins & 24.96 & 29.46 \\
Y3 & \cmark & Ours & \textbf{28.08} & \textbf{31.60} \\
\bottomrule
\end{tabular}
\end{minipage}
\vspace{-0.3cm}
\end{table}

\begin{table}[t]
\tiny
\begin{minipage}[t]{.48\linewidth}
\centering
\caption{Impact of confidence weighting and FNComp on tracking performance.}
\label{tab:quant-ablation-confidence}
\begin{tabular}{c|cc|ccc}
 \toprule
& \textbf{Conf} & \textbf{FNComp} & \textbf{MOTA} & \textbf{IDF1} \\
 \midrule
B1 & \xmark & \xmark & 25.34 & 30.67 \\
B2 & \cmark & \xmark & 28.08 & 31.60 \\
B3 & \cmark & \cmark & \textbf{29.71} & \textbf{35.66} \\
\bottomrule
\end{tabular}
\end{minipage}%
\hfill
\begin{minipage}[t]{.48\linewidth}
\centering
\caption{Impact of quantity of sparse labels on overall model performance.}
\label{tab:quant-ablation-ann-count}
\setlength\tabcolsep{3.7pt}
 \begin{tabular}{c|c|cc}
 \toprule
& \textbf{Anns} & \textbf{MOTA} & \textbf{IDF1} \\
 \midrule
 Q1 & 2 & 18.28 & 23.39 \\
 Q2 & 4 & 28.08 & 31.60 \\
 Q3 & 8 & 28.15 & 32.91 \\
 Q4 & 16 & 32.75 & 37.22 \\
\bottomrule
\end{tabular}
\end{minipage}
\vspace{-0.1cm}
\end{table}

{\parskip=2pt
\noindent\textbf{\textit{Confidence Weighting and FNComp}}: \tabref{tab:quant-ablation-confidence} studies the impact of confidence weighting and the FNComp module on overall tracking performance. We observe that using pseudolabel confidences to model label quality improves MOTA by $2.74$ p.p., while integrating FNComp heatmaps to downweight regions likely affected by false negatives yields a further improvement of $1.63$ p.p. Together, these adaptations increase the IDF1 score by nearly $5$ p.p., with FNComp contributing to approximately $80\%$ of the improvement. These results highlight that explicitly modeling pseudolabel quality and suppressing false negatives significantly enhances tracking accuracy and consistency.
}

{\parskip=2pt
\noindent\textbf{\textit{Sparse Annotation Quantity}}: \tabref{tab:quant-ablation-ann-count} examines the impact of varying the number of per-track annotations on overall tracking performance. As expected, increasing label density results in consistent gains, with MOTA score increasing from $18.28$ p.p. to $32.75$ p.p. when increasing the number of labels per track from two to sixteen. The largest improvement of $9.80$ p.p. occurs when increasing label density from two to four per track, beyond which the gain diminishes, with sixteen annotations yielding only an additional $4.63$ p.p. improvement despite utilizing four times more labels.}

\subsection{Qualitative Results}
\begin{figure}[t]
\centering
\scriptsize
\setlength{\tabcolsep}{0.03cm}
{
\renewcommand{\arraystretch}{0.8}
\newcolumntype{M}[1]{>{\centering\arraybackslash}m{#1}}
\begin{tabular}{M{0.3cm}cM{3.7cm}M{3.7cm}M{3.7cm}}
\\
\multirow{2}{*}{\rotatebox[origin=c]{90}{3D Psuedolabels}} & 
\rotatebox[origin=c]{90}{(a) KITTI} & 
\includegraphics[width=\linewidth, height=1.5cm, frame]{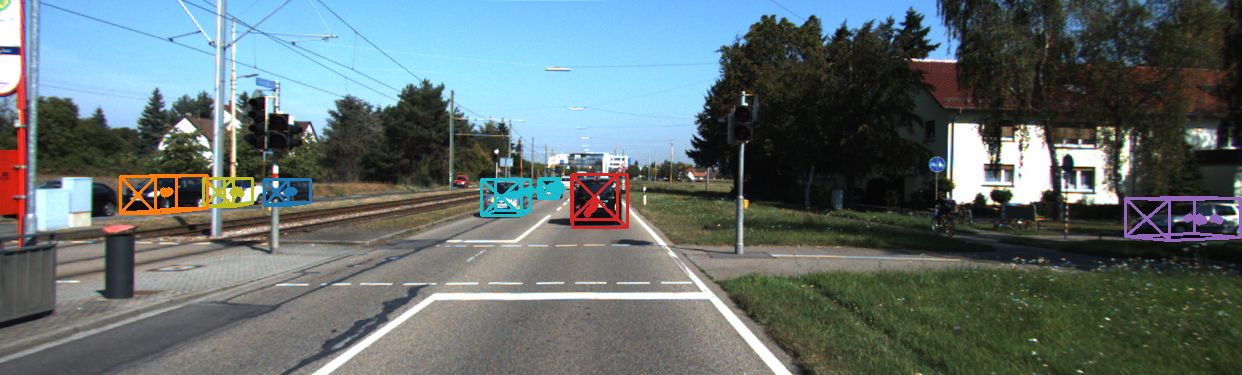} &  \includegraphics[width=\linewidth, height=1.5cm, frame]{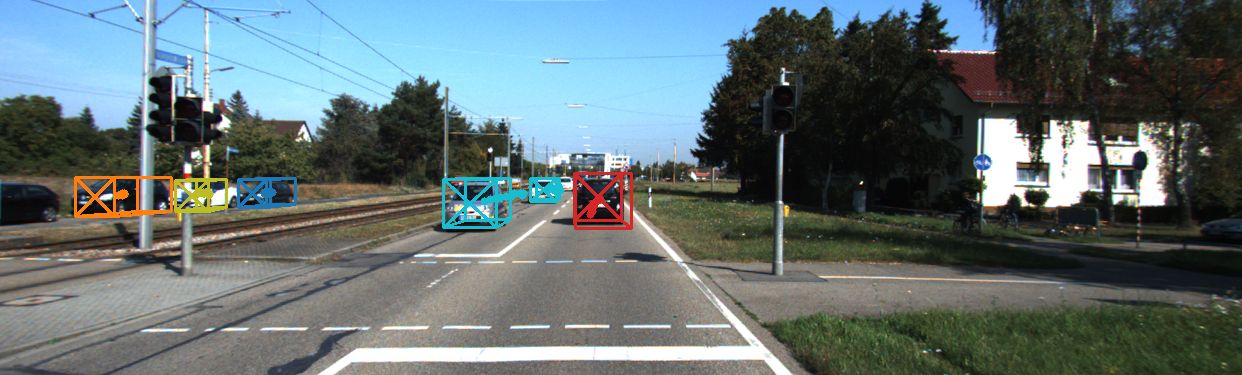} & \includegraphics[width=\linewidth, height=1.5cm, frame]{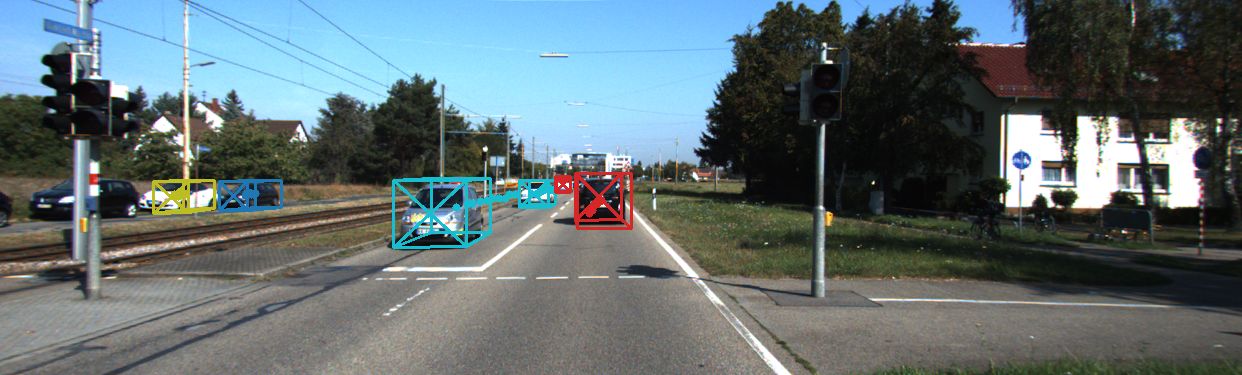}
\\
& \rotatebox[origin=c]{90}{(b) nuScenes} & \includegraphics[width=\linewidth, height=1.5cm, frame]{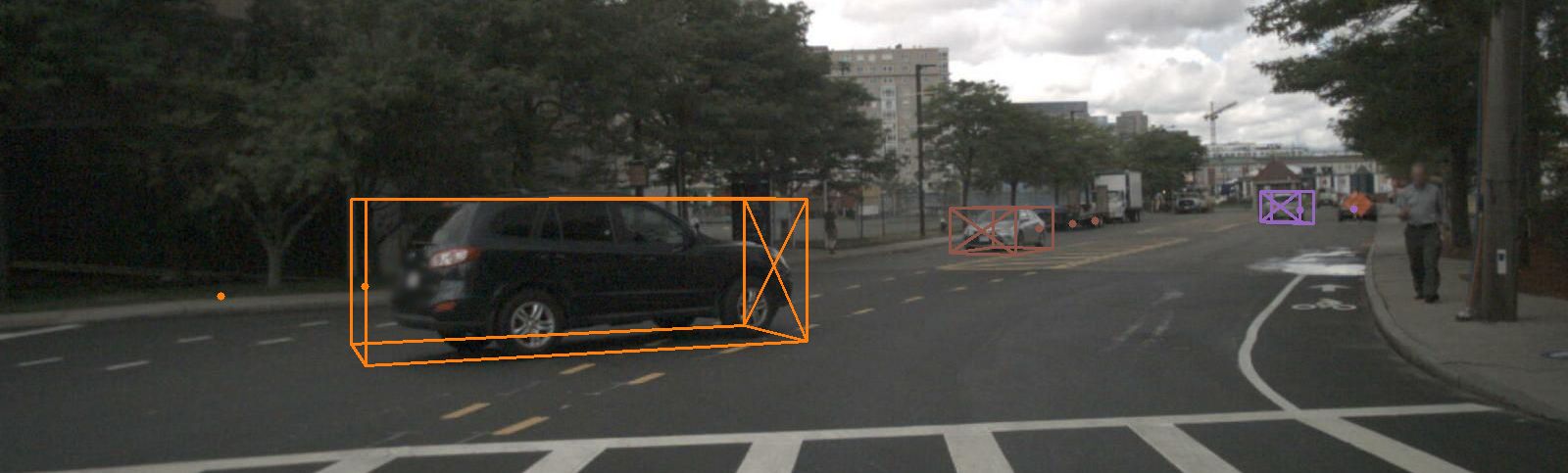} & \includegraphics[width=\linewidth, height=1.5cm, frame]{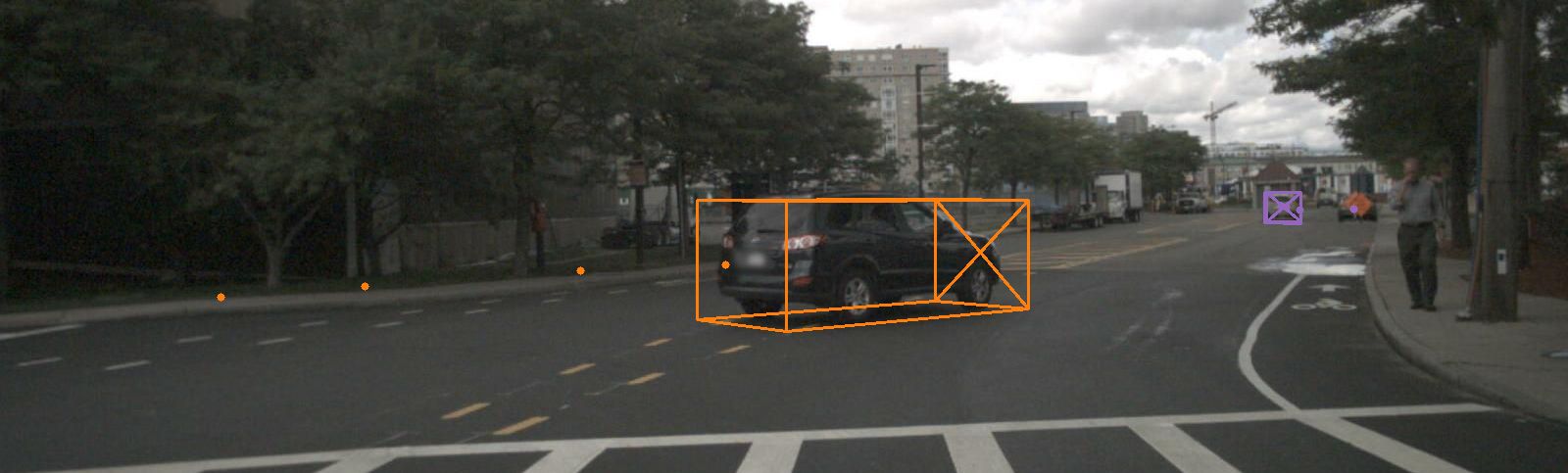} & \includegraphics[width=\linewidth, height=1.5cm, frame]{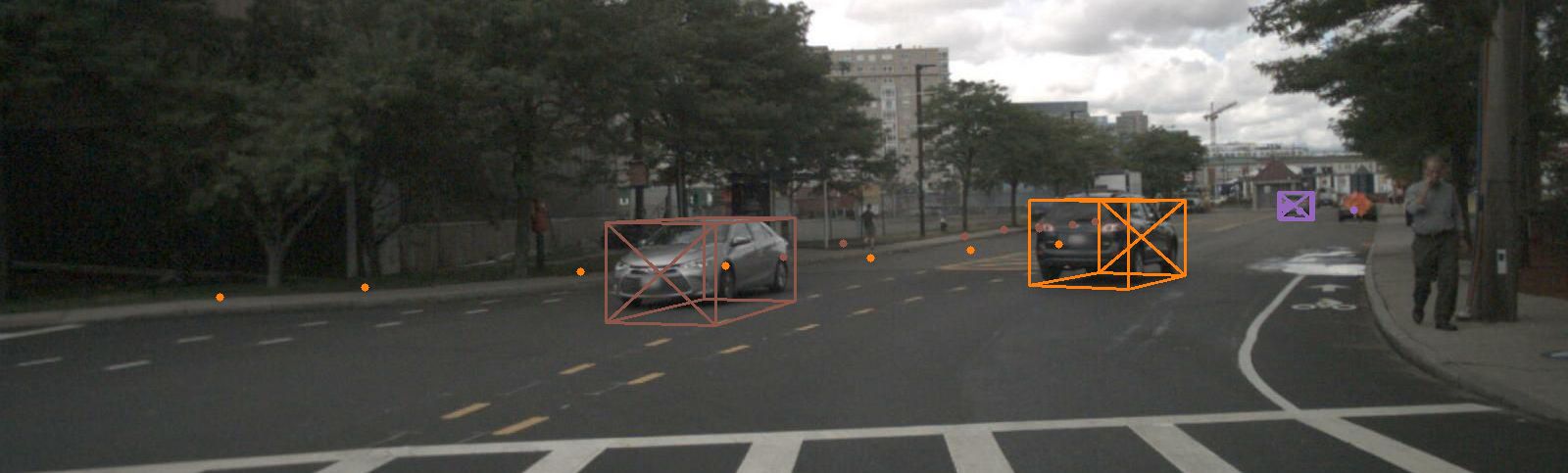}
\\ 
\\
\multirow{2}{*}{\rotatebox[origin=c]{90}{CenterTrack}} & \rotatebox[origin=c]{90}{(c) KITTI} & 
\includegraphics[width=\linewidth, height=1.5cm, frame]{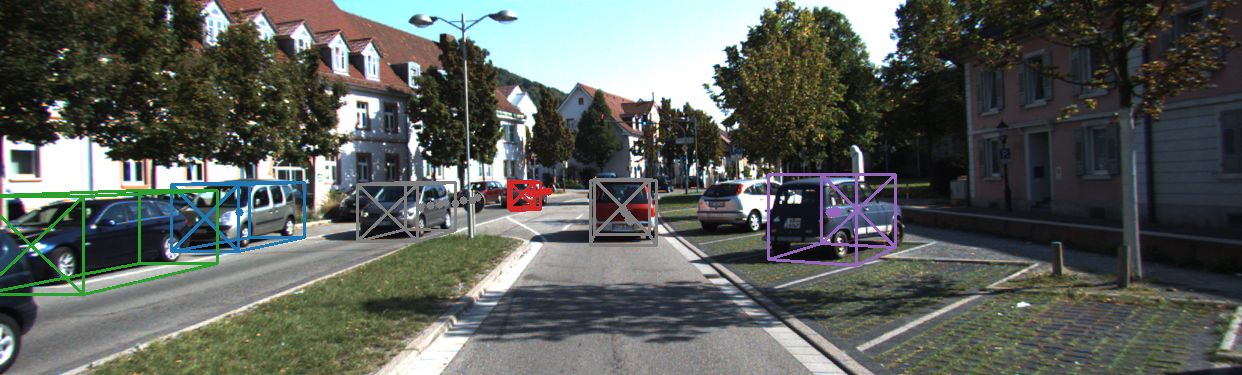} &  \includegraphics[width=\linewidth, height=1.5cm, frame]{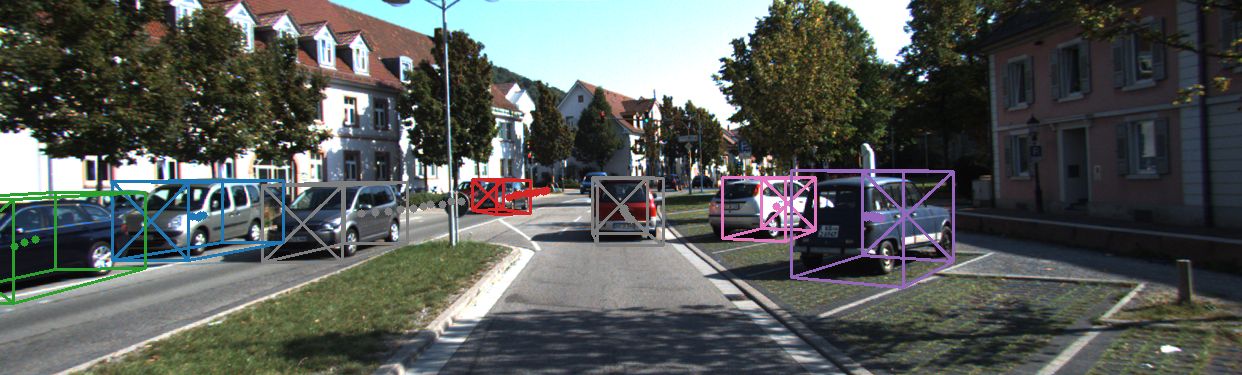} & \includegraphics[width=\linewidth, height=1.5cm, frame]{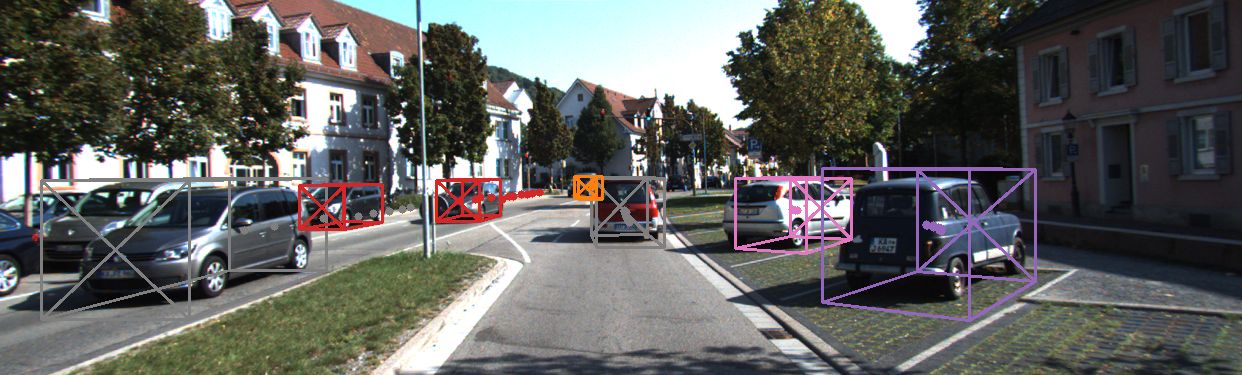}
\\
& \rotatebox[origin=c]{90}{(d) nuScenes} & \includegraphics[width=\linewidth, height=1.5cm, frame]{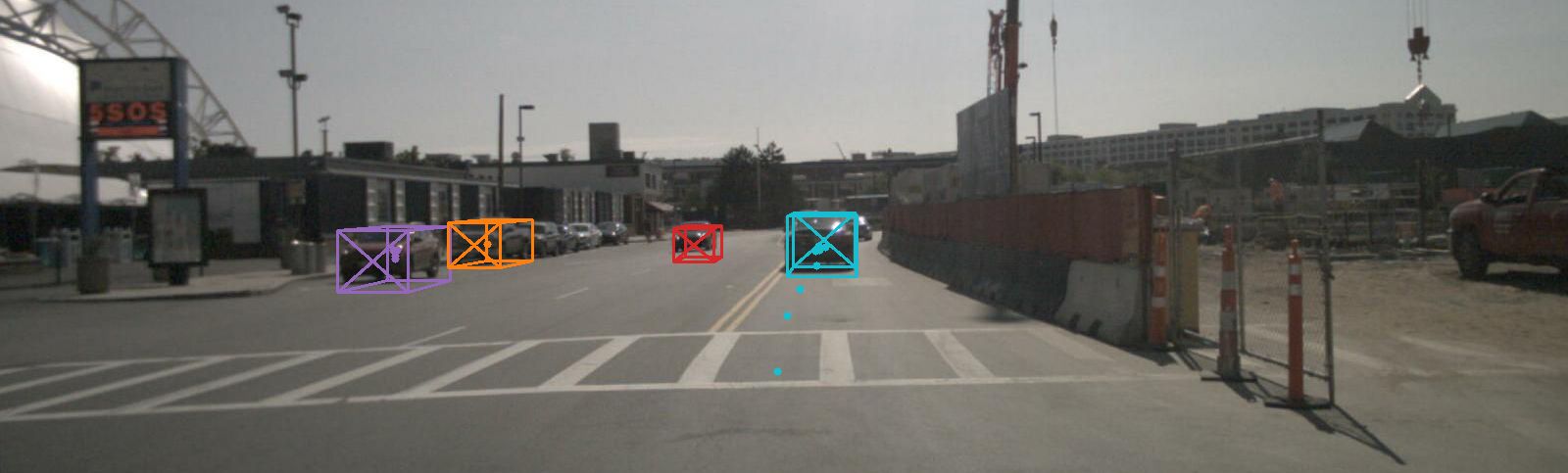} &  \includegraphics[width=\linewidth, height=1.5cm, frame]{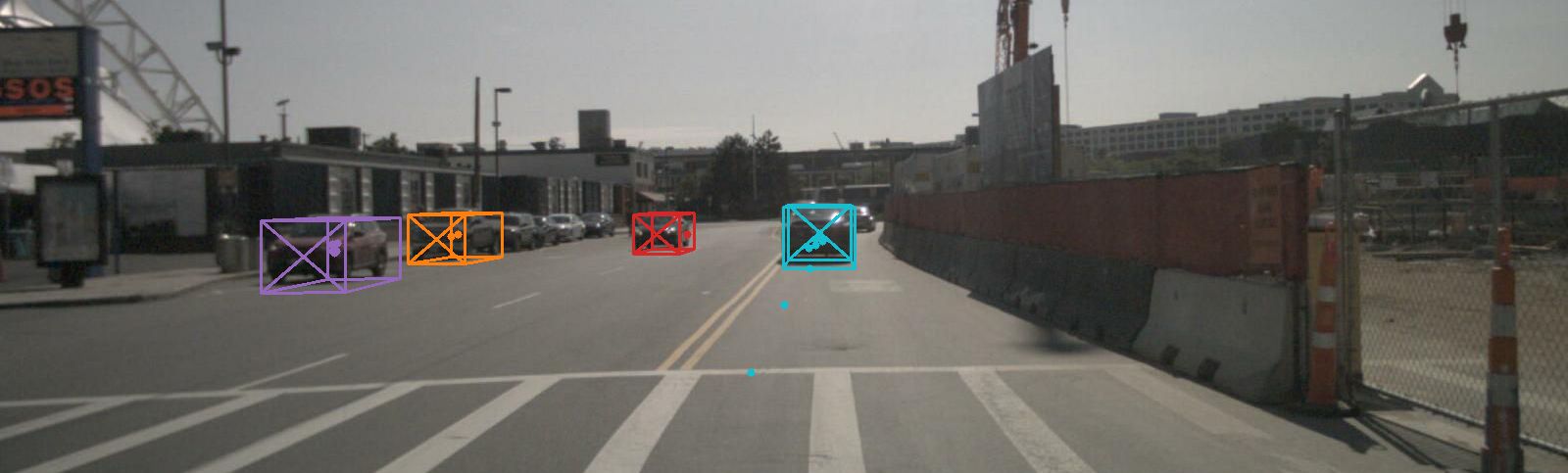} & \includegraphics[width=\linewidth, height=1.5cm, frame]{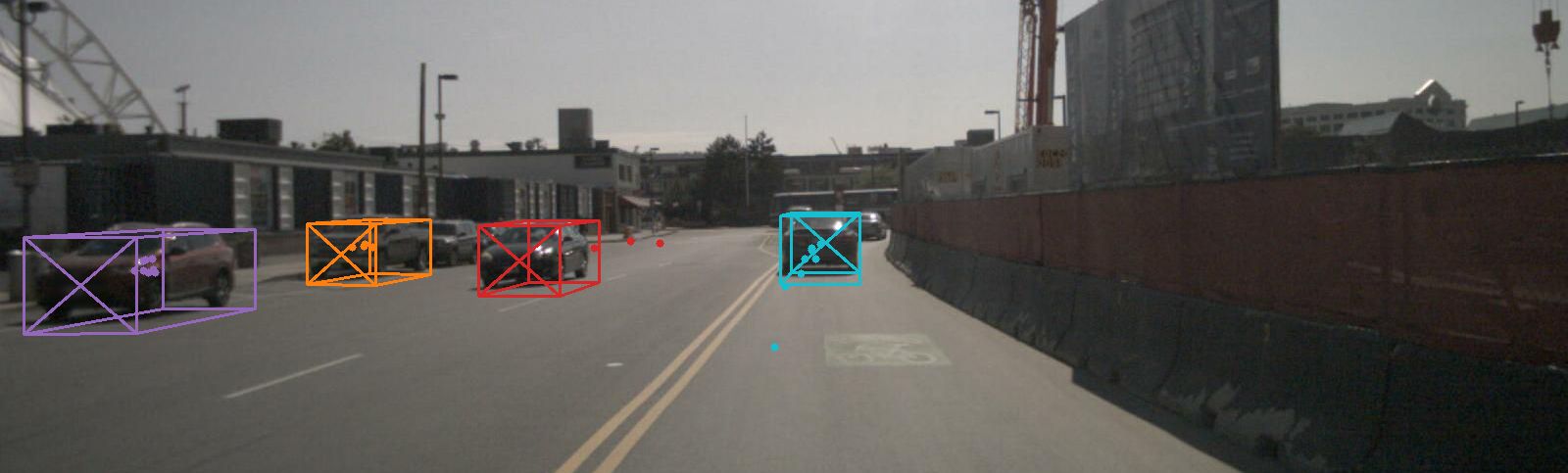}
\end{tabular}
}
\caption{Qualitative results of the 3D pseudolabels generated by~\net~and the corresponding monocular 3D tracking performance when CenterTrack is trained using these pseudolabels. Each object track is visualized in a unique color, and its previous trajectory is illustrated using a sequence of dots of the same color. Note from (a, b) that \net~generates accurate and temporally consistent 3D psuedolabels even when objects undergo total occlusion, allowing CenterTrack to effectively track objects on unseen image sequences (c, d).}
\label{fig:qual-results}
\vspace{-0.1cm}
\end{figure}

We qualitatively evaluate our approach by visualizing both the 3D pseudolabels generated by~\net~as well as the 3D tracking outputs produced by the base network CenterTrack when trained with these pseudolabels. \figref{fig:qual-results} presents these results, where each object track is rendered in a unique color, and its trajectory history is depicted using a sequence of points of the same color.

We observe from \figref{fig:qual-results}(a, b) that our approach accurately localizes objects across frames as well as recovers their 3D pose and dimensions over multiple time steps, thereby producing spatio-temporally consistent 3D tracking pseudolabels. \figref{fig:qual-results}(b) also shows that our framework maintains track consistency through full occlusions, thus demonstrating reliable tracking performance. In \figref{fig:qual-results}(c, d), CenterTrack trained with our 3D pseudolabels effectively generalizes to unseen image sequences and yields stable trajectories and 3D poses that closely align with real-world geometry. 
These results demonstrate that \net~generates accurate and temporally coherent 3D pseudolabels, allowing 3D object tracking frameworks to achieve high-quality performance under sparse supervision. We present additional qualitative results in \secref{sec:supp-qualitative-analysis} of the supplementary material.

\subsection{Discussion of Limitations}
\label{subsec:discussion-of-limitations}

While our approach establishes a new paradigm for sparsely supervised monocular 3D object tracking, it is subject to two main limitations. Firstly, the DINOv2 backbone produces a single feature map of size $\frac{H}{14}$, which can be insufficient for accurately detecting and tracking small and distant objects. Incorporating a multi-scale image encoder could alleviate this limitation by providing richer spatial detail. Secondly, DINOv2 is optimized for objects fully contained within the image frame and thus struggles to produce strong features for objects near image boundaries, which can hinder reliable correspondence estimation. Since these issues relate to an external dependency, i.e., the DINOv2 backbone, they can be resolved by replacing the existing backbone with upcoming newer variants.

%% file: sections/05_conclusion.tex
\section{Conclusion}
\label{sec:conclusion}

In this paper, we presented~\net, the first sparsely supervised framework for monocular 3D object tracking. Our framework decomposes the task into two components, namely, \textit{2D query matching} and \textit{3D geometry estimation}. 2D query matching establishes spatio-temporal correspondence across frames using feature similarity, while 3D geometry estimation recovers accurate 3D object poses by fusing these predictions with self-supervised depth priors. This modular design enables~\net~to generate high-quality 3D pseudolabels, allowing state-of-the-art tracking models to be trained under extreme label sparsity. Extensive experiments with two baseline tracking networks on the KITTI and nuScenes datasets demonstrate that~\net~significantly improves tracking performance, yielding a gain of up to $15.50$ p.p. using only four annotations per track. Future work includes extending this approach to multi-camera setups to enable surround-view 3D object tracking with sparse supervision.

%% file: sections/06_supplementary.tex
\clearpage
\setcounter{page}{1}

\begin{center}
\Large
\textbf{\thetitle}\\
\vspace{0.5em}Supplementary Material \\
\vspace{1.0em}
\end{center}

\setcounter{section}{0}
\setcounter{equation}{0}
\setcounter{figure}{0}
\setcounter{table}{0}
\makeatletter

\renewcommand{\thesection}{S.\arabic{section}}
\renewcommand{\thesubsection}{S.\arabic{section}.\arabic{subsection}}
\renewcommand{\thetable}{S.\arabic{table}}
\renewcommand{\thefigure}{S.\arabic{figure}}

In this supplementary material, we present further results on \net~for sparsely supervised monocular 3D object tracking. Specifically, we analyze the tracking performance on DEFT~\cite{cit:track_mot3d-deft} in \secref{sec:supp-quantitate-results}, and present additional qualitative results on both the KITTI and nuScenes datasets in \secref{sec:supp-qualitative-analysis}.

\section{Quantitative Results}
\label{sec:supp-quantitate-results}
In this section, we report the 3D tracking performance obtained when using DEFT~\cite{cit:track_mot3d-deft} as the base tracking model, as summarized in \tabref{tab:quant-pseudolabel-eval} of the main paper.
As shown in \tabref{tab:quant-pseudolabel-eval}, \net~significantly improves upon the sparse supervision baseline, outperforming it by $14.73$ p.p. on the KITTI dataset and $5.96$ p.p. on the nuScenes dataset. This improvement demonstrates the effectiveness of our 3D pseudolabel generation framework in producing consistent and reliable training samples for downstream model training.
We further observe that our approach consistently outperforms SPOT-based baselines on the KITTI and nuScenes datasets, highlighting the strengths of the 2D query matching and 3D geometry estimation modules in generating accurate 3D tracking pseudolabels. Additionally, we observe that our approach outperforms both SAM2-based baselines on the KITTI dataset by nearly $4$ p.p., despite relying only on sparse annotations rather than the millions of labeled samples used to train the SAM2 foundation model. On the nuScenes dataset, our approach remains within $1.08$ p.p. of the SAM2 baseline, further highlighting its data efficiency. We also observe that combining our 3D geometry estimation module with SAM2 yields the strongest overall performance, outperforming all baselines by more than $4$ p.p. on the KITTI dataset and nearly $10$ p.p. on the nuScenes dataset. These results show that our 3D geometry estimation module can effectively leverage depth cues from a self-supervised depth network to produce accurate geometry predictions, even with sparse labels. Together, these results demonstrate that the 3D pseudolabels generated by \net~enable superior monocular 3D object tracking across different base networks and datasets, thereby confirming the broad applicability of our framework.

\section{Additional Qualitative Results}
\label{sec:supp-qualitative-analysis}

In this section, we present additional qualitative results on the KITTI and nuScenes datasets for both the 3D pseudolabels generated by \net~as well as the 3D tracking predictions generated by the base network, CenterTrack, when trained using the generated 3D pseudolabels.

\begin{figure}[t]
\centering
\scriptsize
\setlength{\tabcolsep}{0.03cm}
{
\renewcommand{\arraystretch}{0.8}
\newcolumntype{M}[1]{>{\centering\arraybackslash}m{#1}}
\begin{tabular}{M{0.3cm}cM{3.7cm}M{3.7cm}M{3.7cm}}
\\
\multirow{15}{*}{\rotatebox[origin=c]{90}{KITTI}} & (a) & 
\includegraphics[width=\linewidth, height=1.5cm, frame]{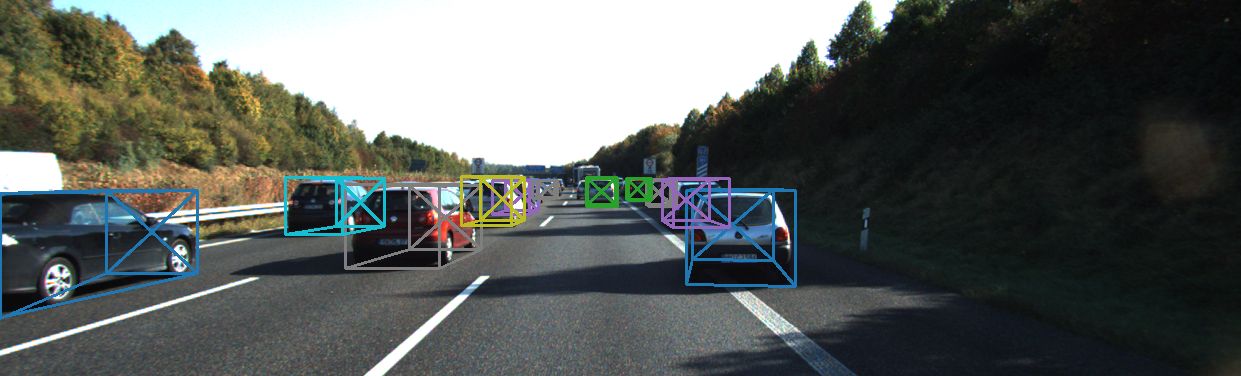} &  \includegraphics[width=\linewidth, height=1.5cm, frame]{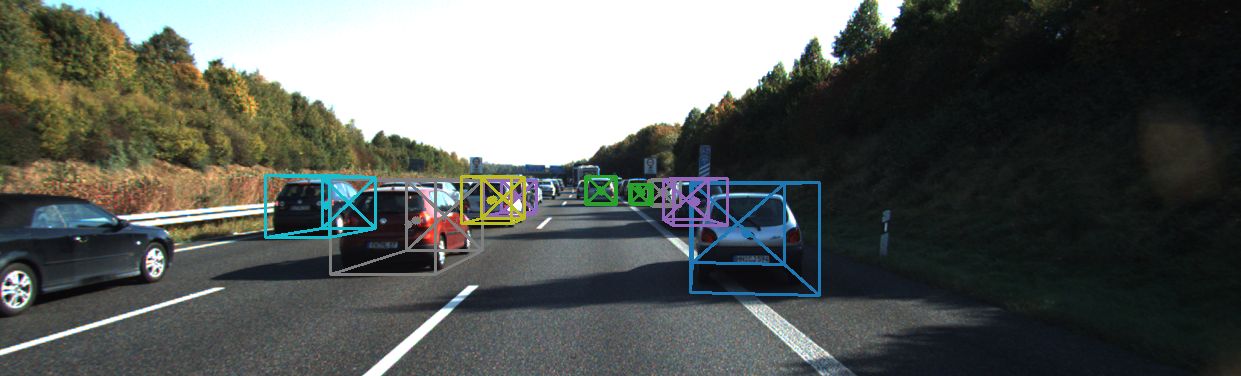} & \includegraphics[width=\linewidth, height=1.5cm, frame]{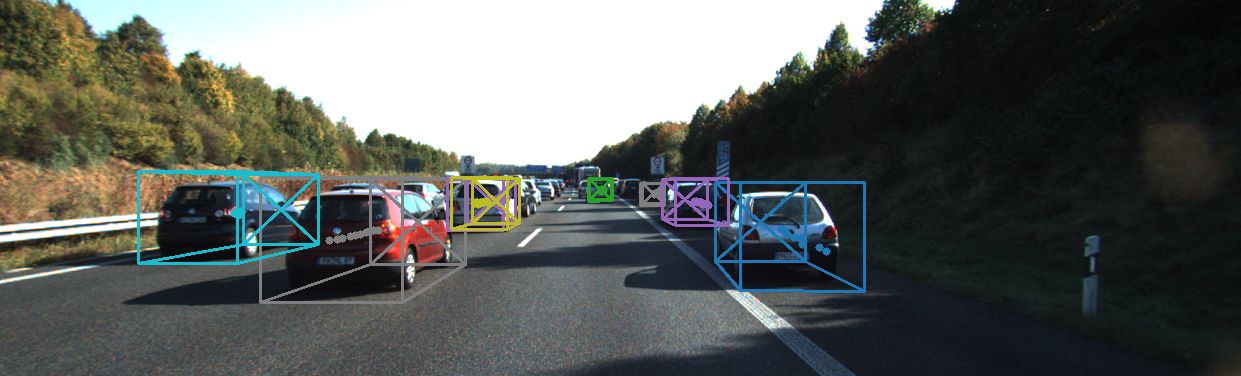}
\\
& (b) & 
\includegraphics[width=\linewidth, height=1.5cm, frame]{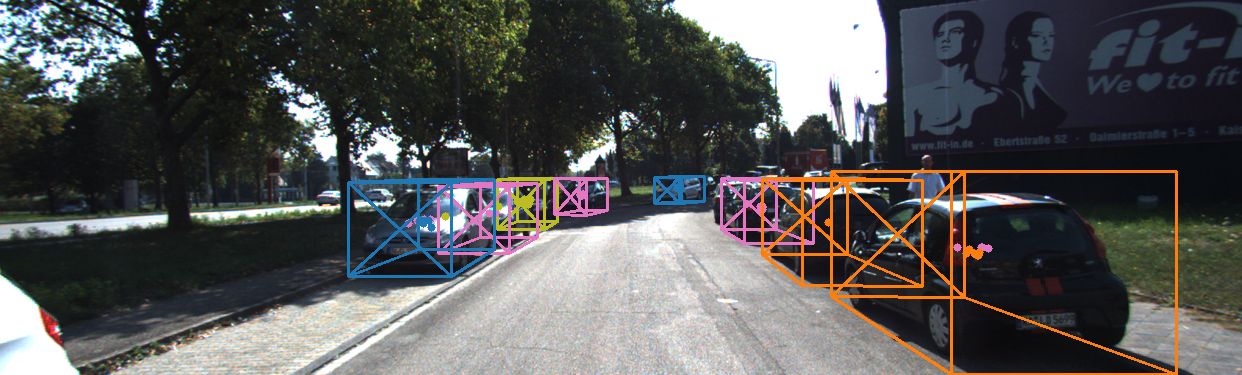} &  \includegraphics[width=\linewidth, height=1.5cm, frame]{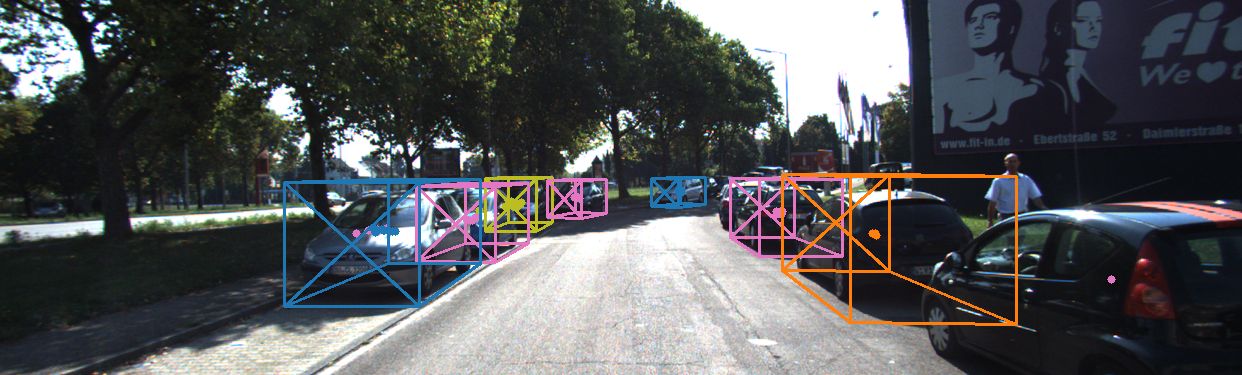} & \includegraphics[width=\linewidth, height=1.5cm, frame]{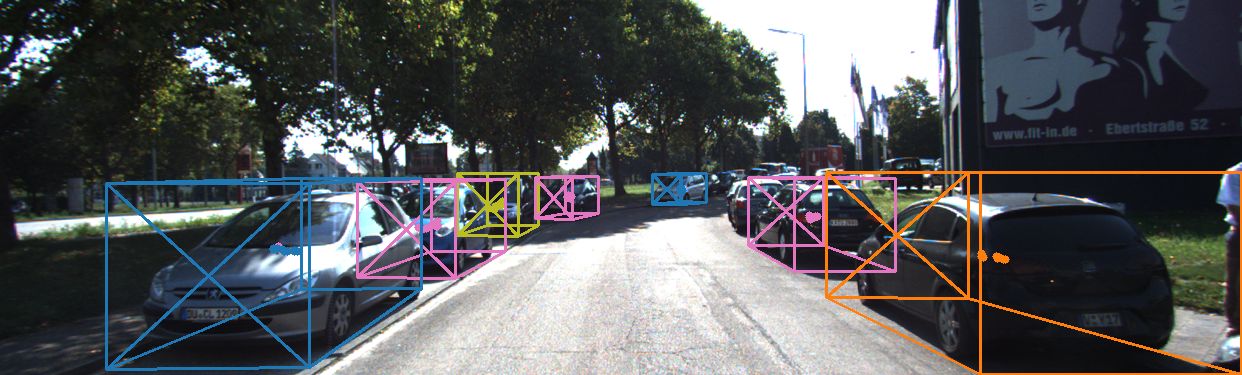}
\\
& (c) & 
\includegraphics[width=\linewidth, height=1.5cm, frame]{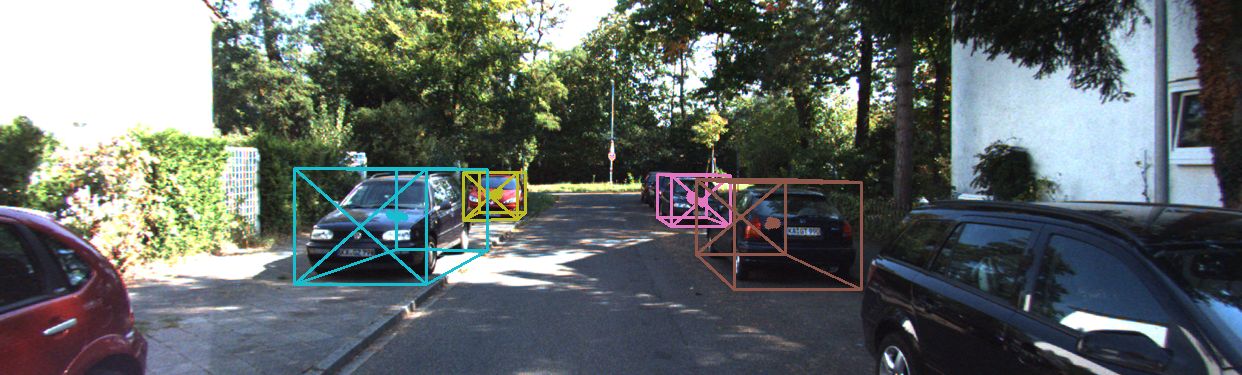} &  \includegraphics[width=\linewidth, height=1.5cm, frame]{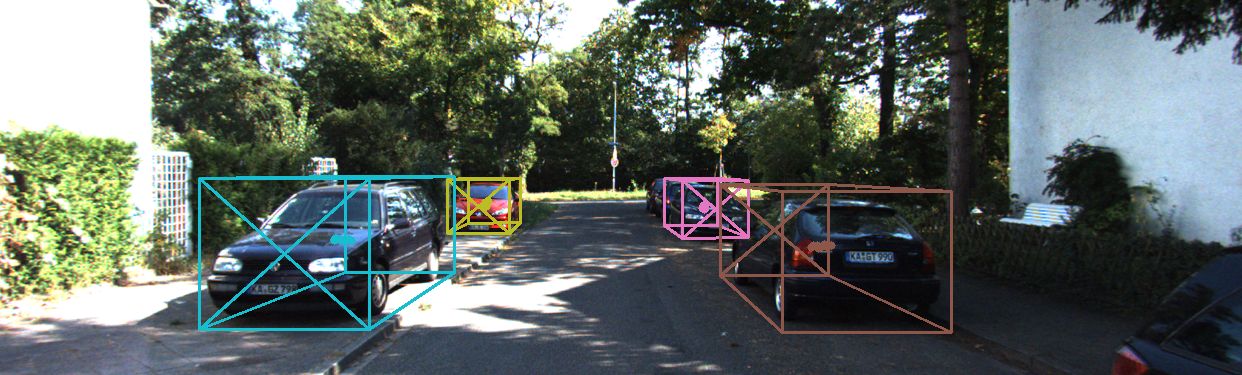} & \includegraphics[width=\linewidth, height=1.5cm, frame]{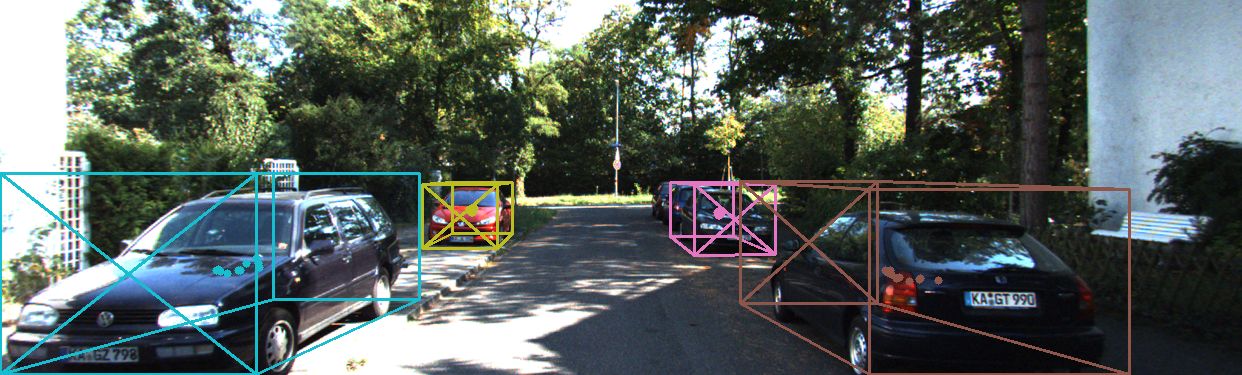}
\\
\\
\multirow{15}{*}{\rotatebox[origin=c]{90}{nuScenes}} & (d) & 
\includegraphics[width=\linewidth, height=1.5cm, frame]{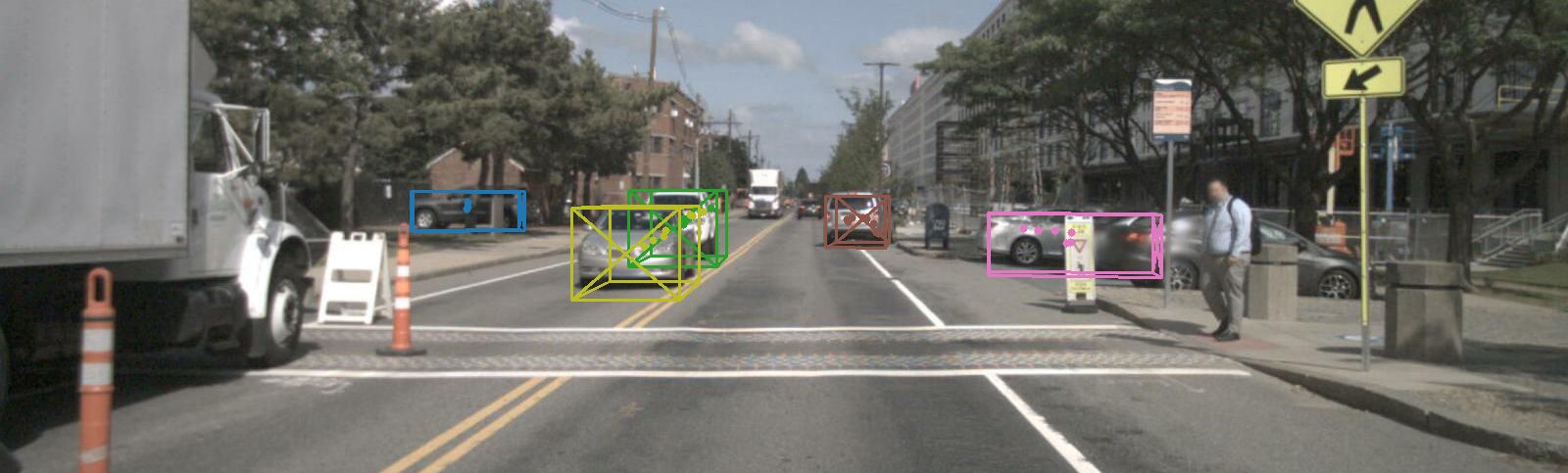} & \includegraphics[width=\linewidth, height=1.5cm, frame]{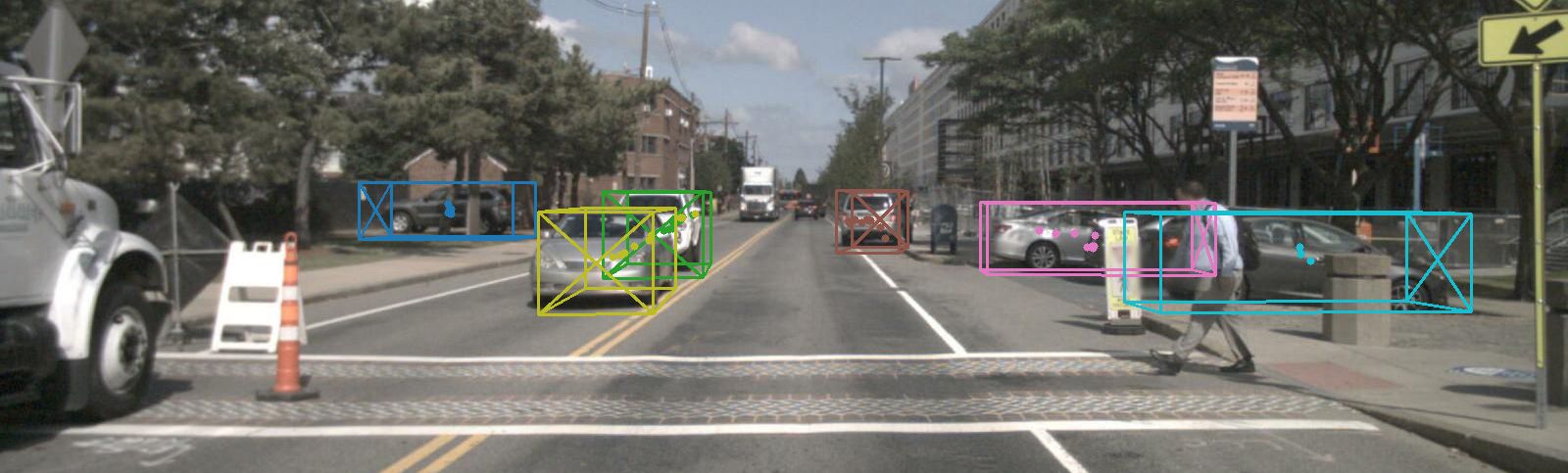} & \includegraphics[width=\linewidth, height=1.5cm, frame]{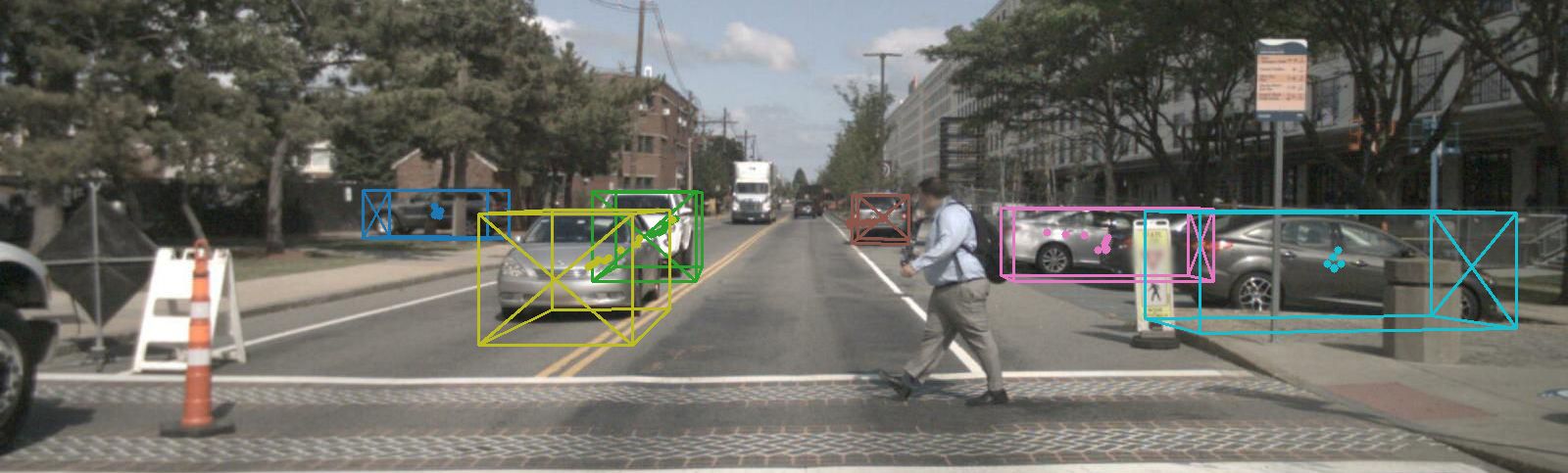}
\\
& (e) & 
\includegraphics[width=\linewidth, height=1.5cm, frame]{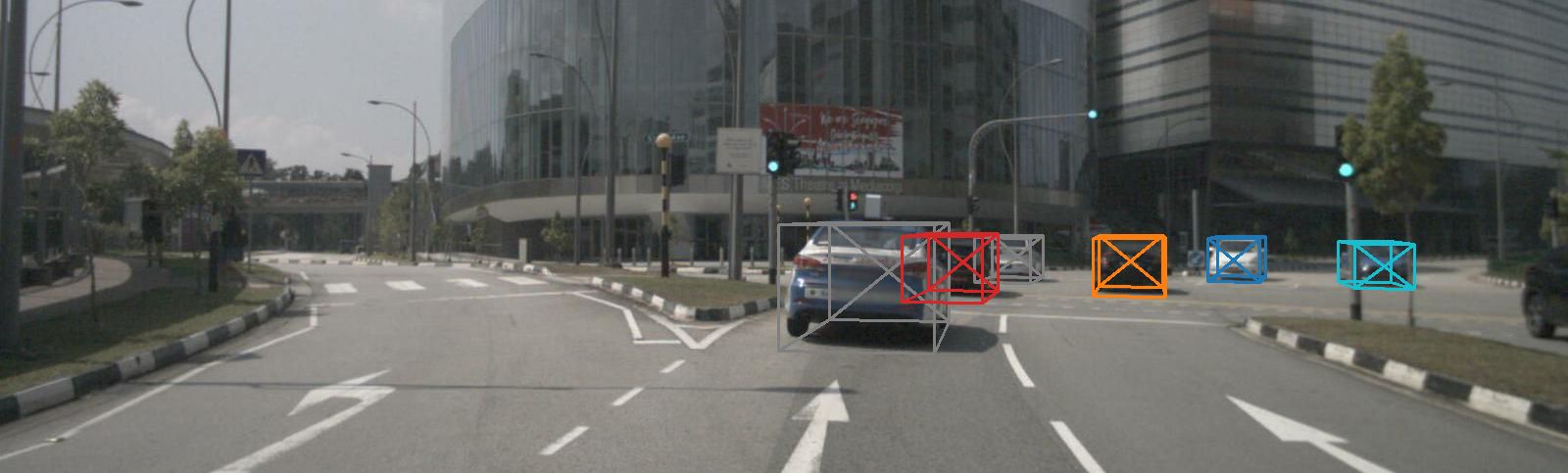} & \includegraphics[width=\linewidth, height=1.5cm, frame]{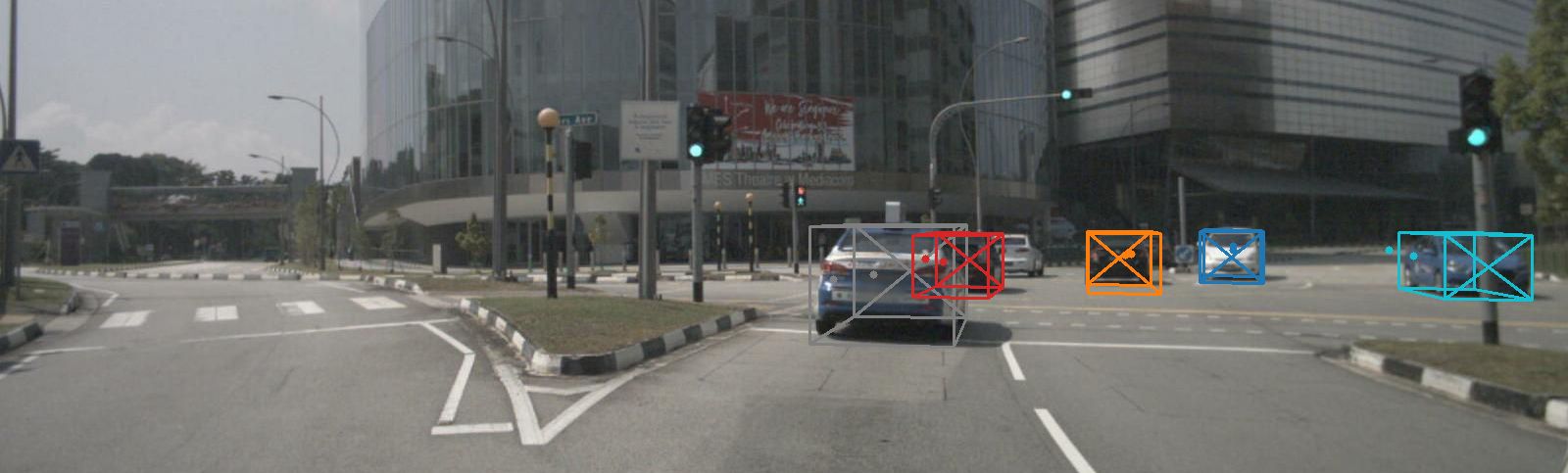} & \includegraphics[width=\linewidth, height=1.5cm, frame]{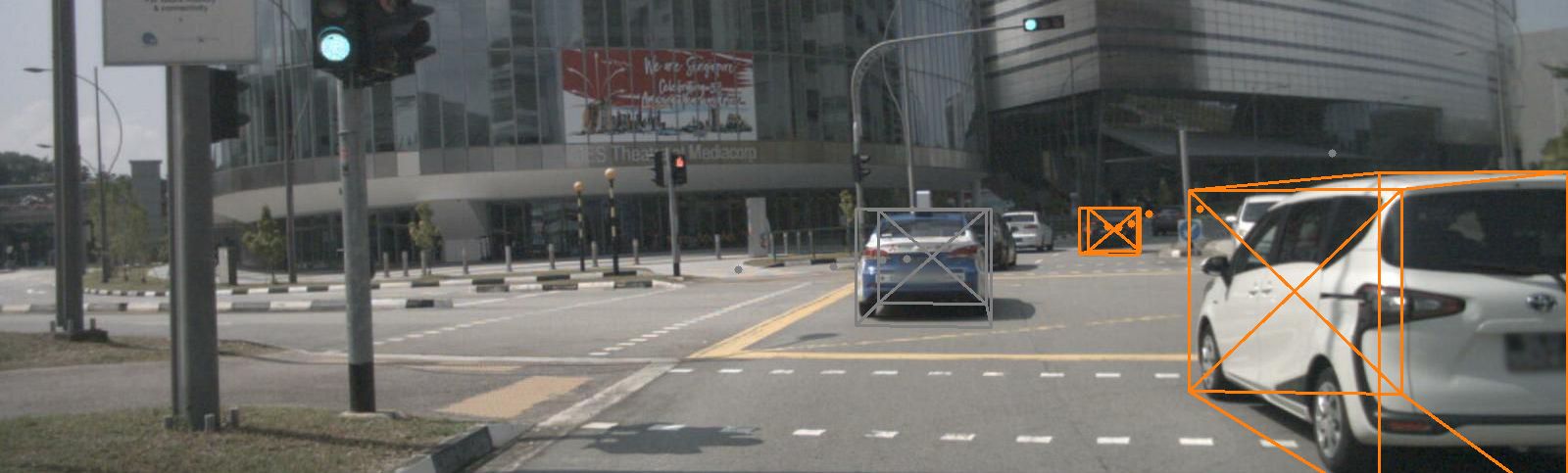}
\\
 & (f) & 
\includegraphics[width=\linewidth, height=1.5cm, frame]{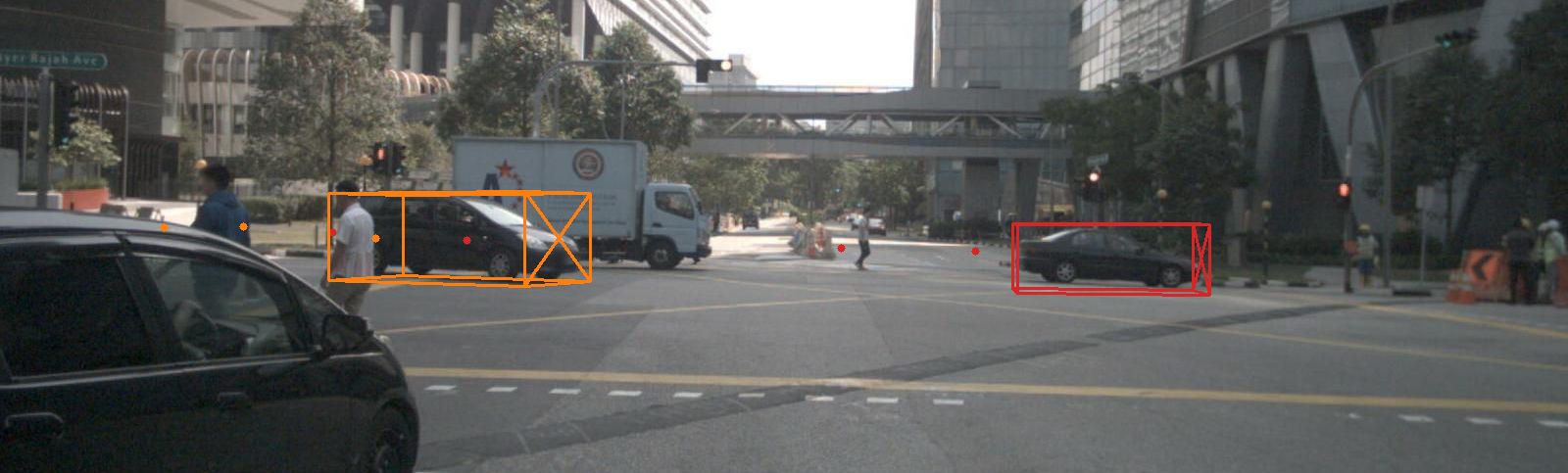} & \includegraphics[width=\linewidth, height=1.5cm, frame]{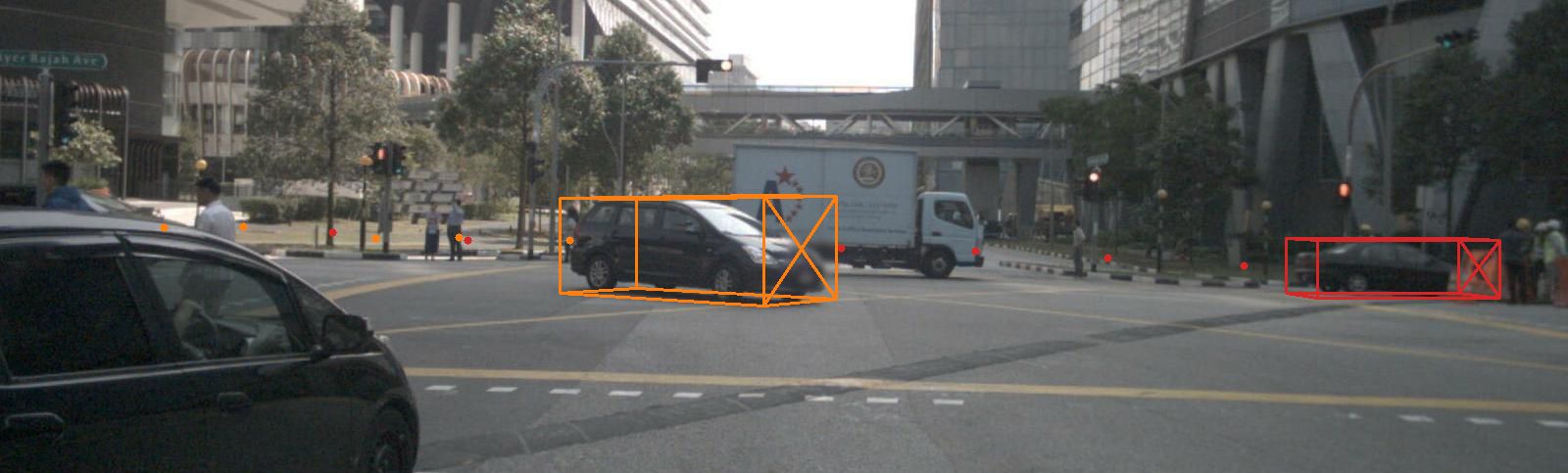} & \includegraphics[width=\linewidth, height=1.5cm, frame]{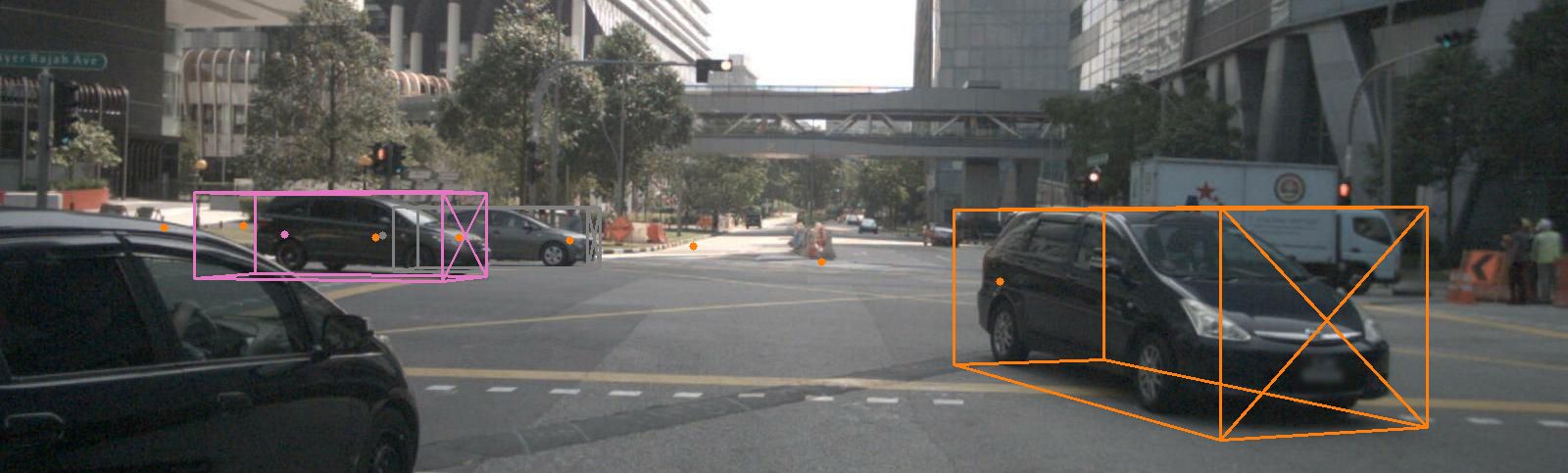}
\\
\end{tabular}
}
\caption{Visualization of the 3D pseudolabels generated by \net. Each object track is visualized in a unique color, and its previous trajectory is illustrated using a sequence of dots of the same color. Note the accurate and temporally consistent predictions across both the KITTI and nuScenes datasets.}
\label{fig:supp-qual-plabels-good}
\vspace{-0.3cm}
\end{figure}

\subsection{3D Pseudolabels}
\label{subsec:supp-qual-3d-pseudolabels}
We demonstrate the quality of our 3D pseudolabels by projecting them onto multiple image sequences from both the KITTI and nuScenes datasets. \figref{fig:supp-qual-plabels-good} presents these results. 
As illustrated in \figref{fig:supp-qual-plabels-good}(a, b, c), \net~produces temporally stable and geometrically consistent 3D pseudolabels even under challenging conditions such as the presence of multiple visually similar objects, heavy clutter, and significant occlusions. This robustness is largely enabled by our similarity estimation module, which accurately identifies correspondences between the query object and target frames, thereby promoting track continuity.
Further, \figref{fig:supp-qual-plabels-good}(c) highlights a scenario where our approach correctly estimates the orientation of a vehicle in an atypical configuration. The pink bounding box denotes a car parked on the right side of the road but facing the ego vehicle. Despite this uncommon layout, the predicted yaw remains accurate, underscoring the benefit of our yaw estimation module in such view geometries.

\figref{fig:supp-qual-plabels-good}(d, e, f) highlight cross-domain generalization where our approach generates consistent pseudolabels for image sequences from geographically different regions in the USA and Singapore. This shows that our approach is not tightly coupled to region-specific information and can generalize well across diverse domains. Additionally, these image sequences show that our model maintains spatially consistent 3D pose estimates for multiple objects across time steps, highlighting its temporal coherence. We also observe a failure mode in \figref{fig:supp-qual-plabels-good}(f) where the black car next to the ego vehicle is not tracked by our model. As discussed in~\secref{subsec:discussion-of-limitations}, this problem stems from the DINOv2 backbone producing weaker features for objects near image boundaries, thereby reducing matching confidence. However, our FNComp module is designed to detect such false negatives and mitigate their effect during base network training.

\begin{figure}[t]
\centering
\scriptsize
\setlength{\tabcolsep}{0.03cm}
{
\renewcommand{\arraystretch}{0.8}
\newcolumntype{M}[1]{>{\centering\arraybackslash}m{#1}}
\begin{tabular}{M{0.3cm}cM{3.7cm}M{3.7cm}M{3.7cm}}
\\
\multirow{8}{*}{\rotatebox[origin=c]{90}{KITTI}} & (a) & 
\includegraphics[width=\linewidth, height=1.5cm, frame]{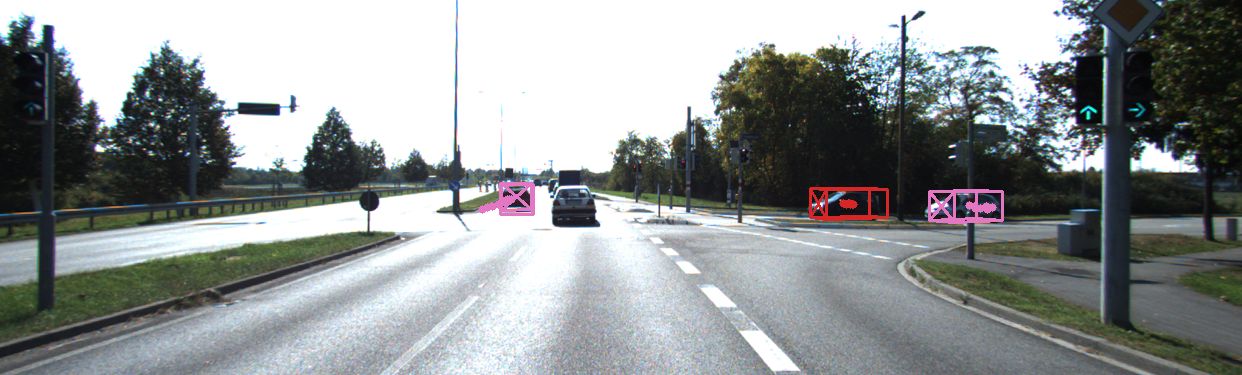} &  \includegraphics[width=\linewidth, height=1.5cm, frame]{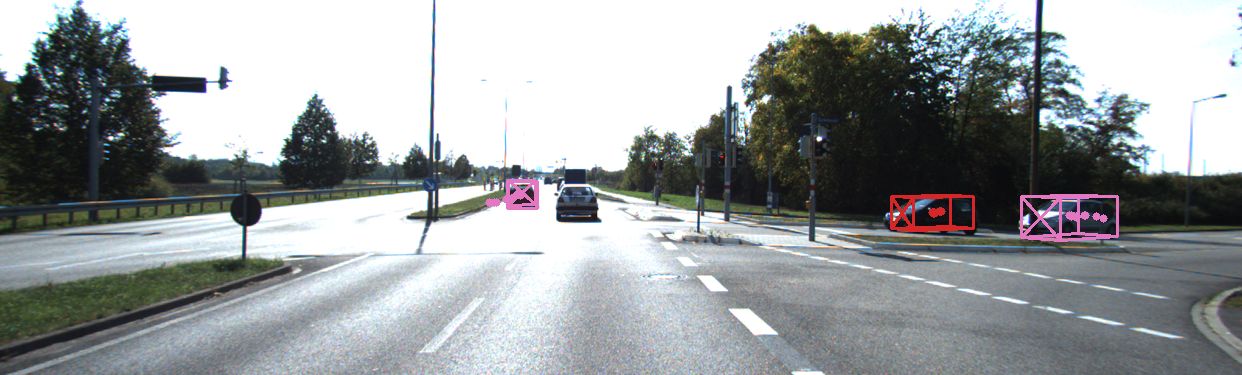} & \includegraphics[width=\linewidth, height=1.5cm, frame]{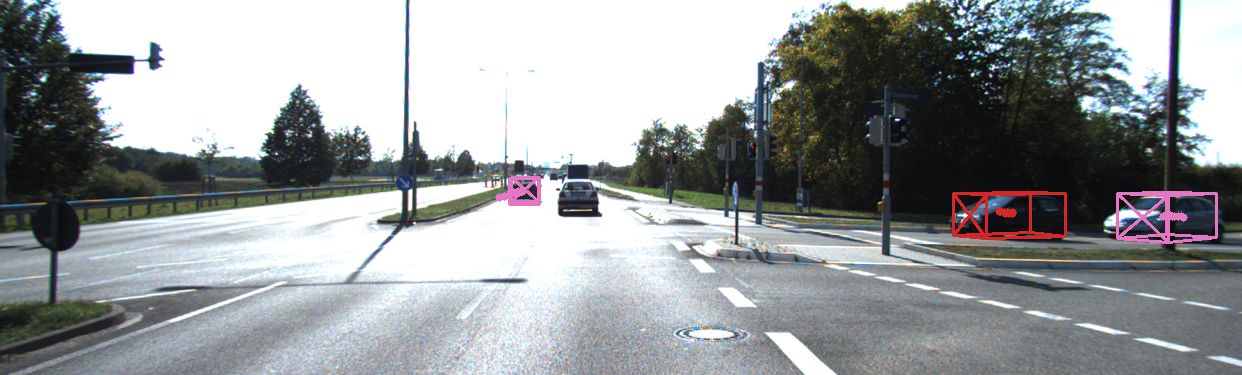}
\\
 & (b) & 
\includegraphics[width=\linewidth, height=1.5cm, frame]{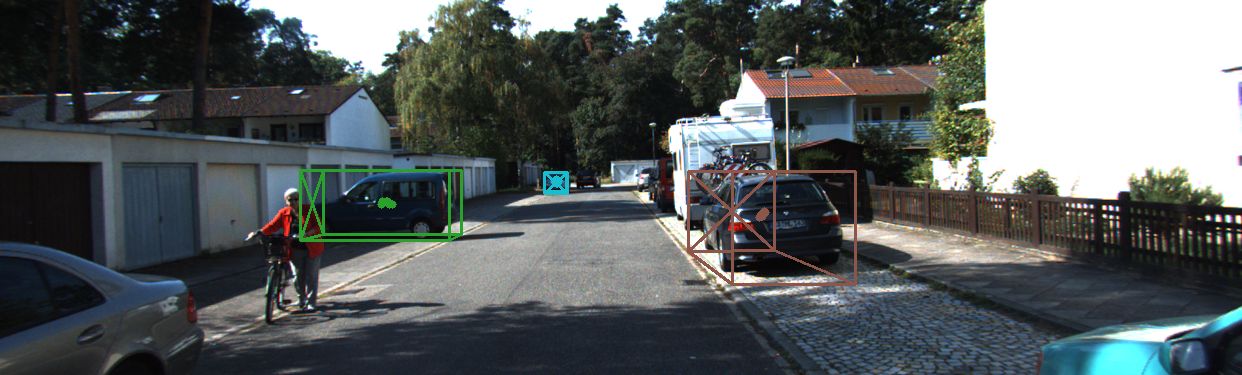} &  \includegraphics[width=\linewidth, height=1.5cm, frame]{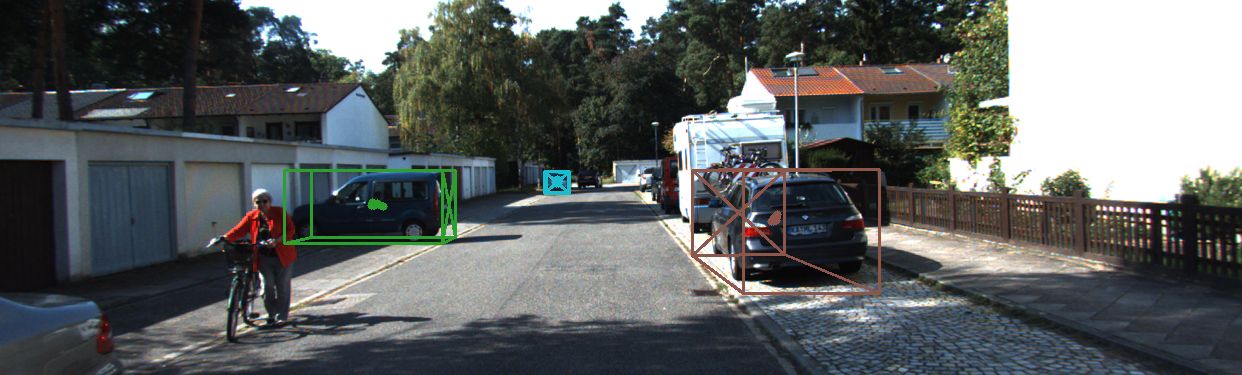} & \includegraphics[width=\linewidth, height=1.5cm, frame]{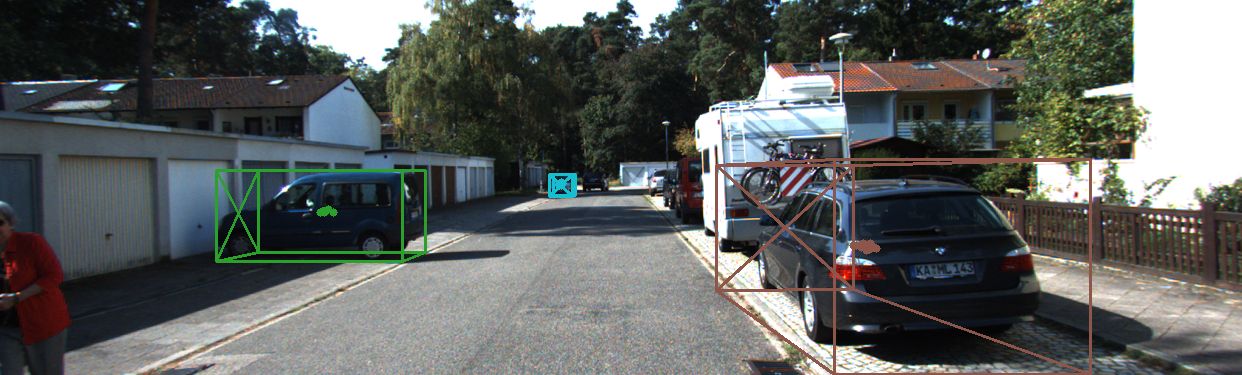}
\\
\\
\multirow{8}{*}{\rotatebox[origin=c]{90}{nuScenes}} & (c) & 
\includegraphics[width=\linewidth, height=1.5cm, frame]{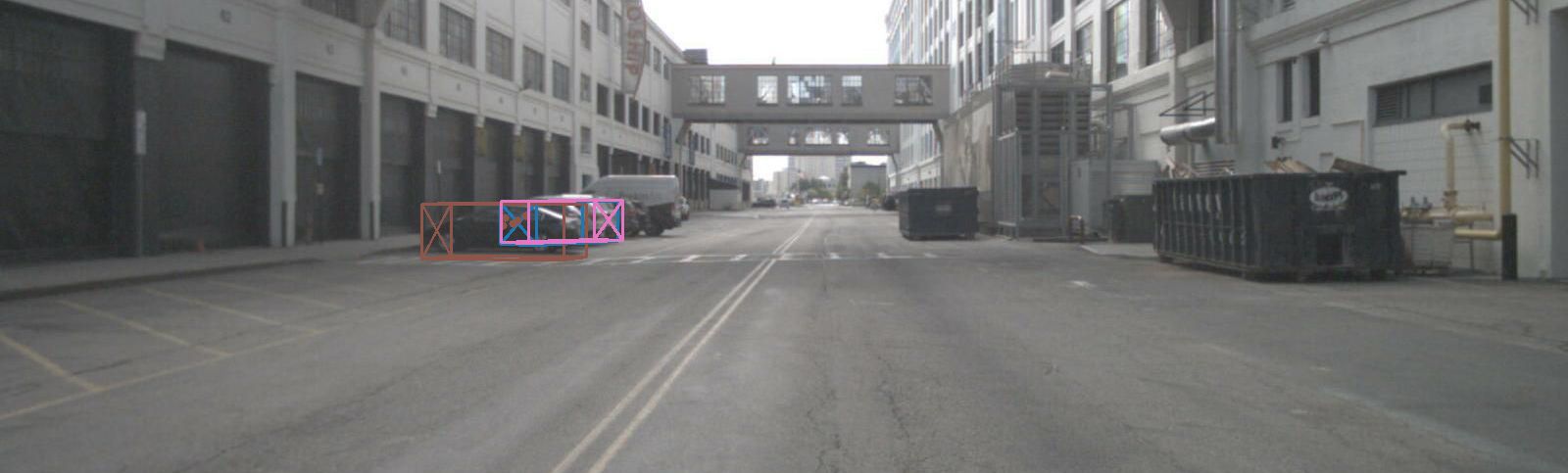} &  \includegraphics[width=\linewidth, height=1.5cm, frame]{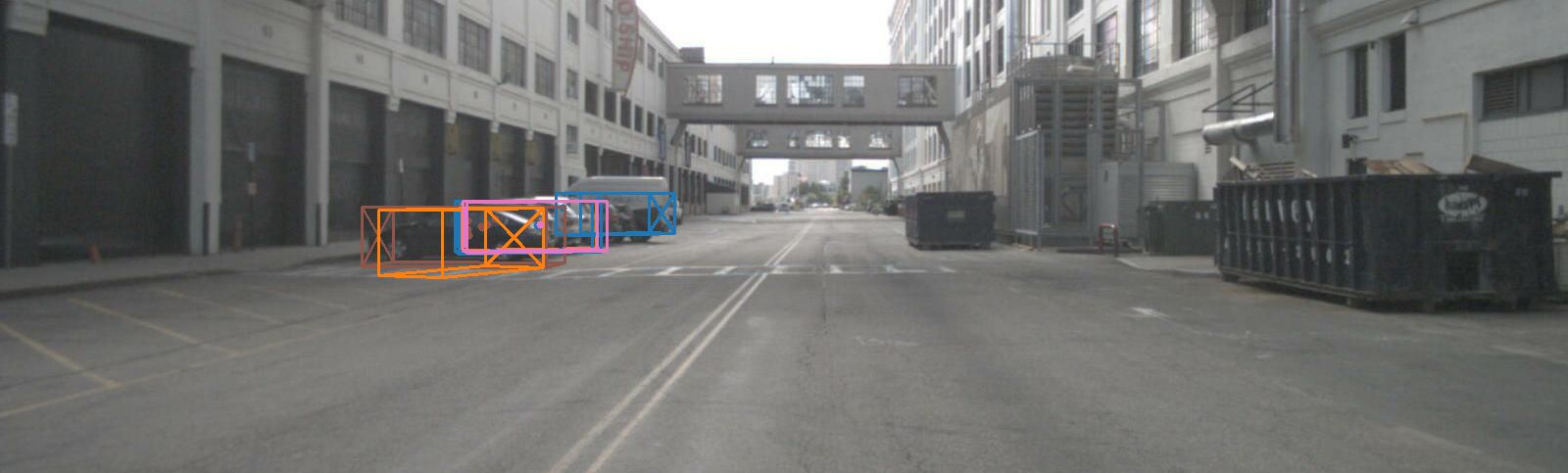} & \includegraphics[width=\linewidth, height=1.5cm, frame]{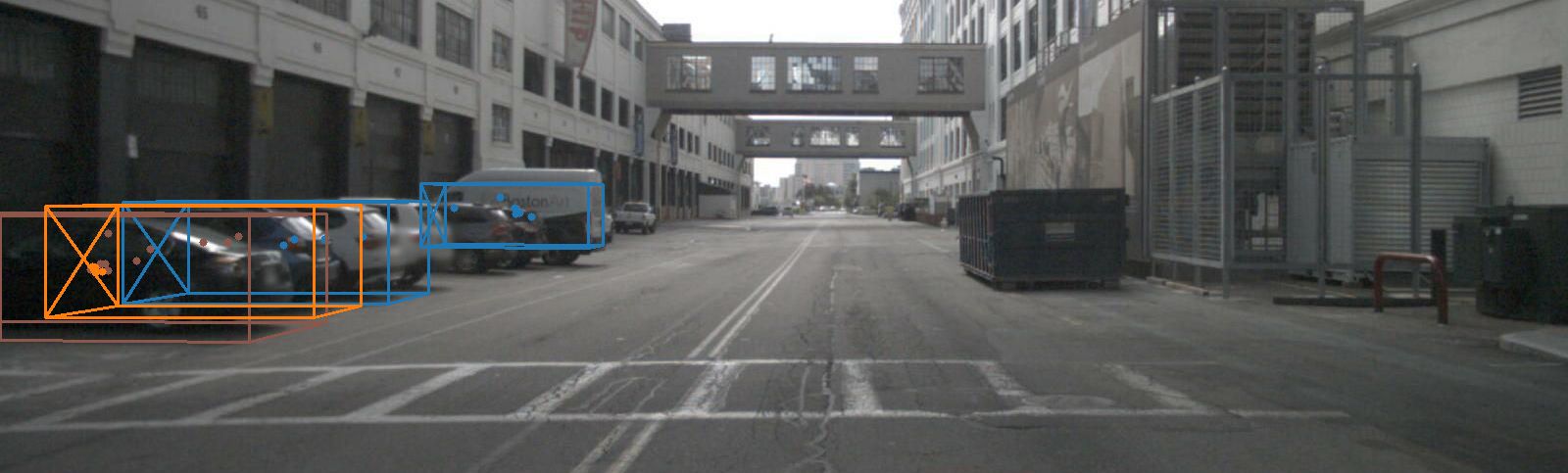}
\\
& (d) & 
\includegraphics[width=\linewidth, height=1.5cm, frame]{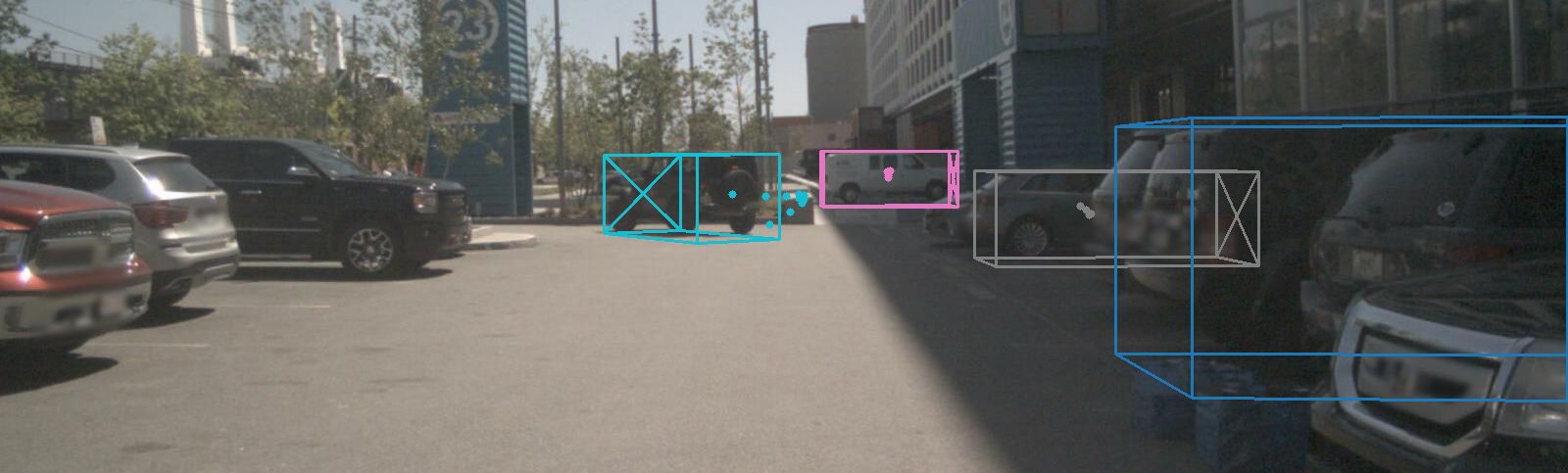} &  \includegraphics[width=\linewidth, height=1.5cm, frame]{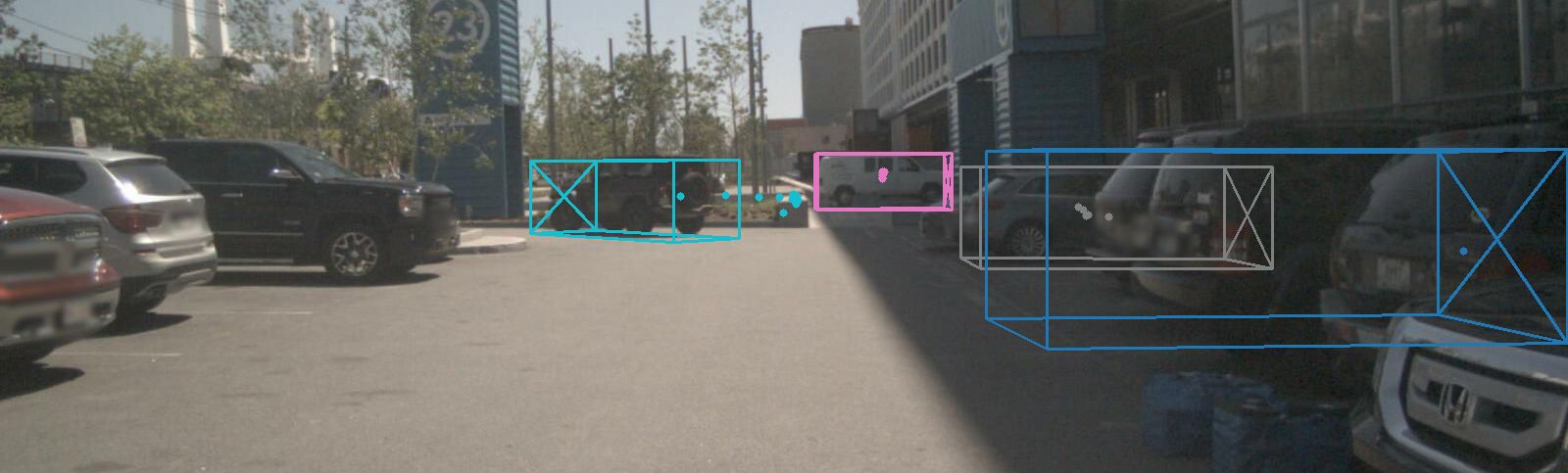} & \includegraphics[width=\linewidth, height=1.5cm, frame]{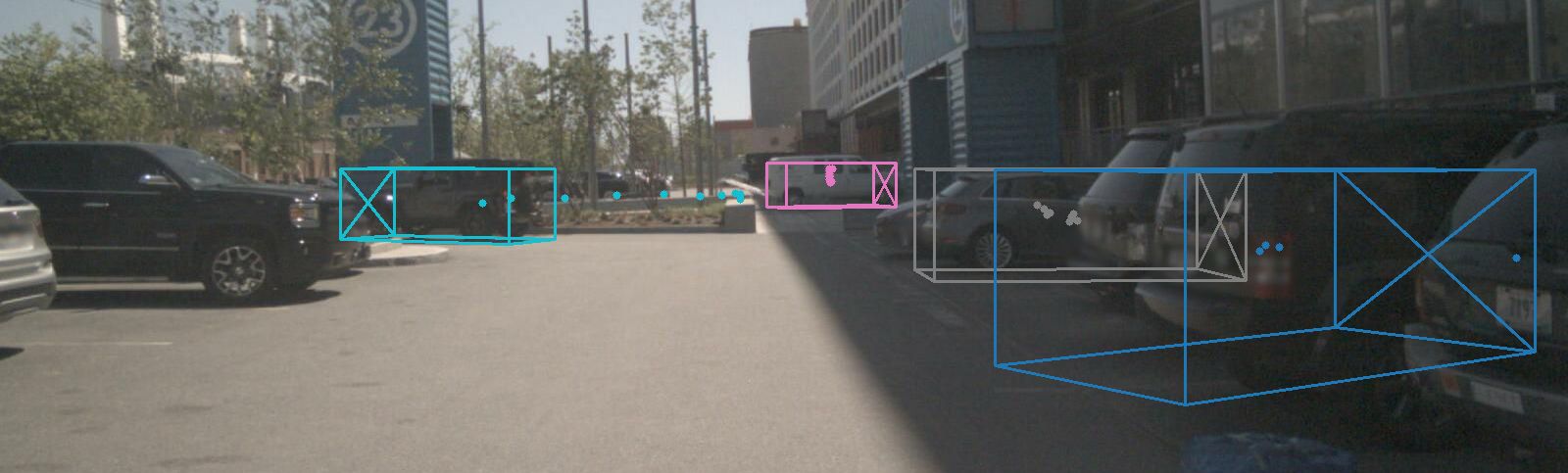} \\

\end{tabular}
}
\caption{Visualization of the failure modes experienced by \net. Each object track is visualized in a unique color, and its previous trajectory is illustrated using a sequence of dots of the same color. Observe that failures typically occur when the target object is far away, near image edges, or subject to significant occlusion and object clutter.}
\label{fig:supp-qual-plabel-fail}
\vspace{-0.3cm}
\end{figure}

\subsubsection{Additional Failures:}
We present additional failure cases in \figref{fig:supp-qual-plabel-fail} for both the KITTI and nuScenes datasets. We observe from \figref{fig:supp-qual-plabel-fail}(a) that our model sometimes fails to track small and distant objects in the scene. This limitation primarily stems from the use of a DINOv2 backbone, which downsamples input images by a factor of $14$, suppressing fine-grained information necessary for reliable association and resulting in premature track termination. In \figref{fig:supp-qual-plabel-fail}(b), the orientation of the vehicle highlighted in green flips by 180 degrees across the image sequence. This ambiguity is caused by the direction vector $d$, which fails to determine whether the object is facing towards or away from the ego vehicle. This problem is further amplified by the near-perpendicular orientation of the car relative to the ego vehicle, making it extremely challenging to determine its true orientation. Finally, \figref{fig:supp-qual-plabel-fail}(c, d) shows instances from the nuScenes dataset where the 2D query matching module struggles to identify correct object associations. Such erroneous matching is observed in image sequences with severe, sustained occlusions and multiple vehicles with highly similar appearances. These factors hinder the generation of discriminative visual features and in turn affect the 2D matching performance.

\begin{figure}[t]
\scriptsize
\setlength{\tabcolsep}{0.03cm}
{
\renewcommand{\arraystretch}{0.8}
\newcolumntype{M}[1]{>{\centering\arraybackslash}m{#1}}
\begin{tabular}{M{0.3cm}cM{3.7cm}M{3.7cm}M{3.7cm}}
\\
\multirow{15}{*}{\rotatebox[origin=c]{90}{KITTI}} & (a) & 
\includegraphics[width=\linewidth, height=1.5cm, frame]{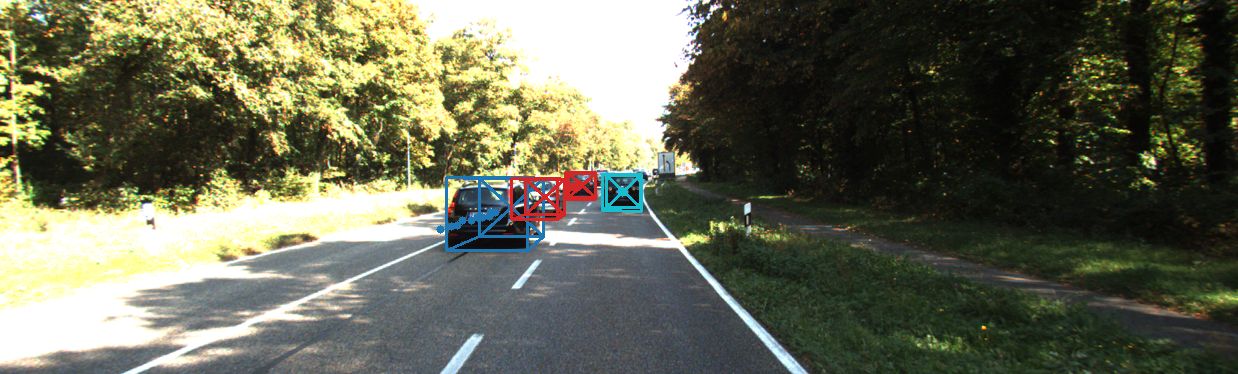} &  \includegraphics[width=\linewidth, height=1.5cm, frame]{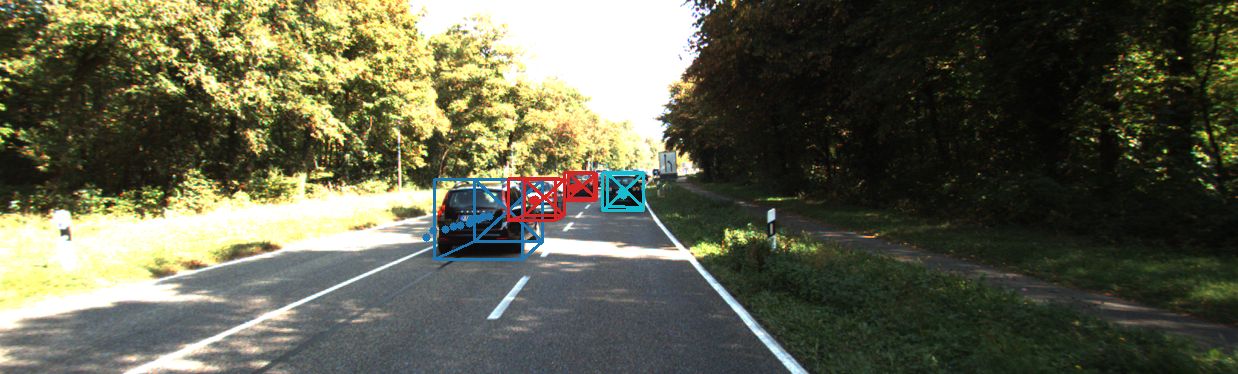} & \includegraphics[width=\linewidth, height=1.5cm, frame]{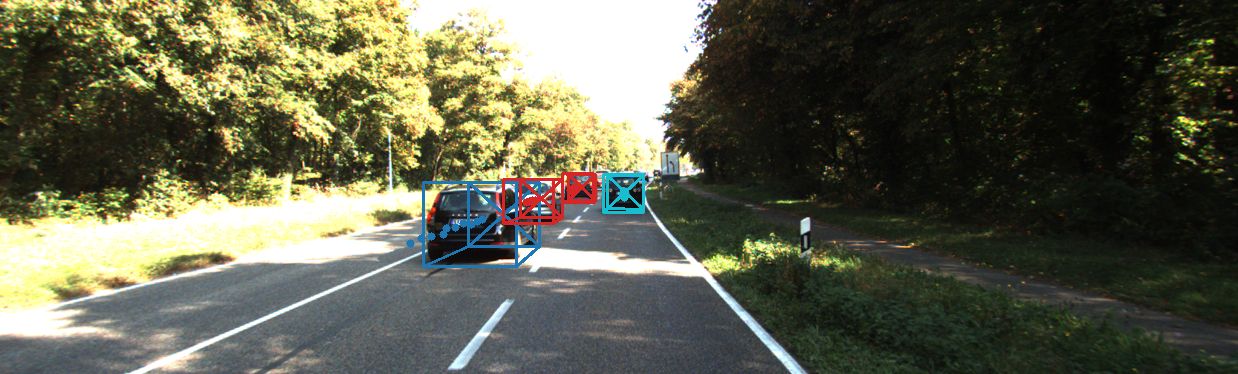}
\\
 & (b) & 
\includegraphics[width=\linewidth, height=1.5cm, frame]{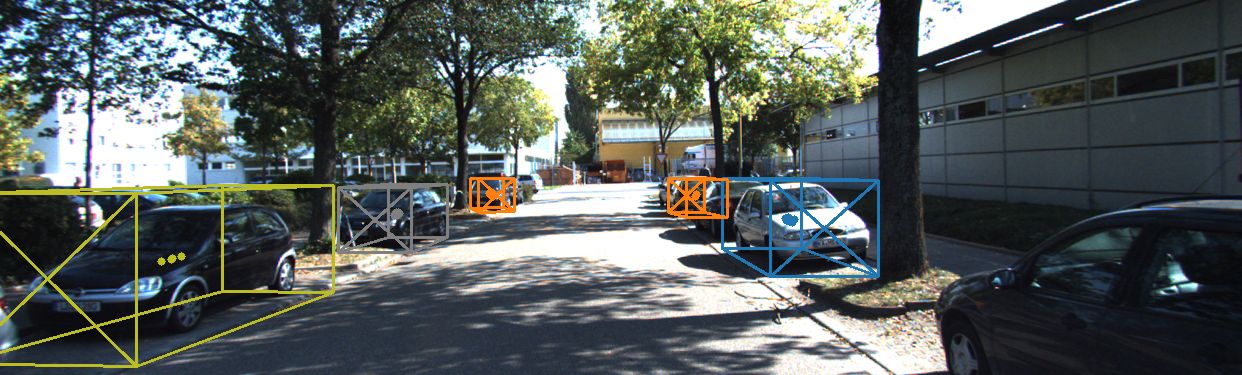} &  \includegraphics[width=\linewidth, height=1.5cm, frame]{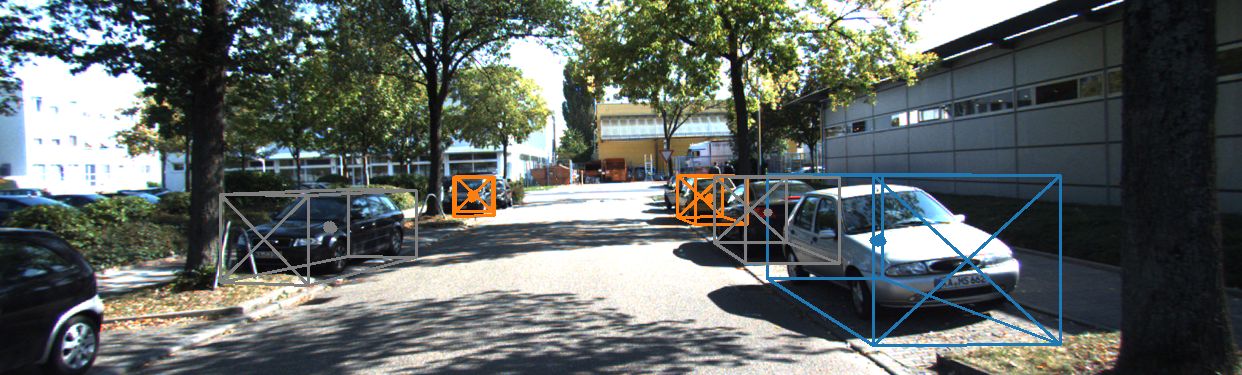} & \includegraphics[width=\linewidth, height=1.5cm, frame]{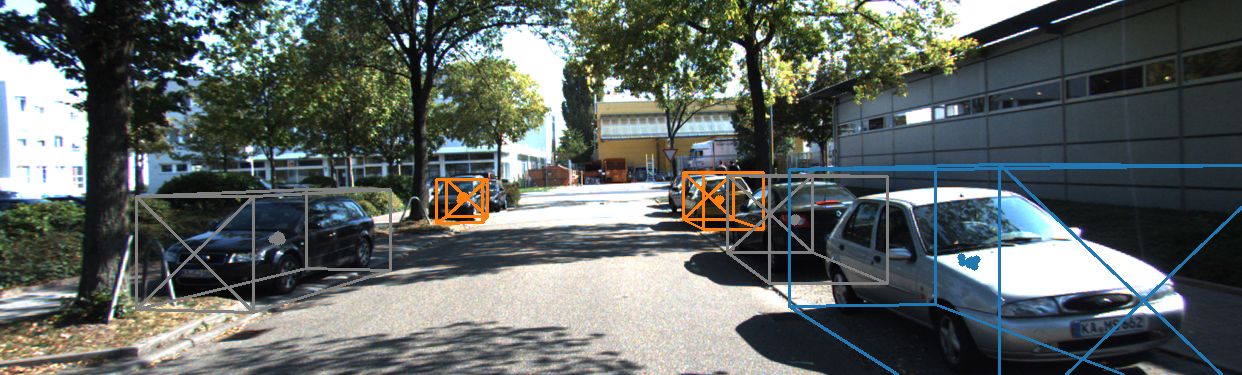}
\\
& (c) & 
\includegraphics[width=\linewidth, height=1.5cm, frame]{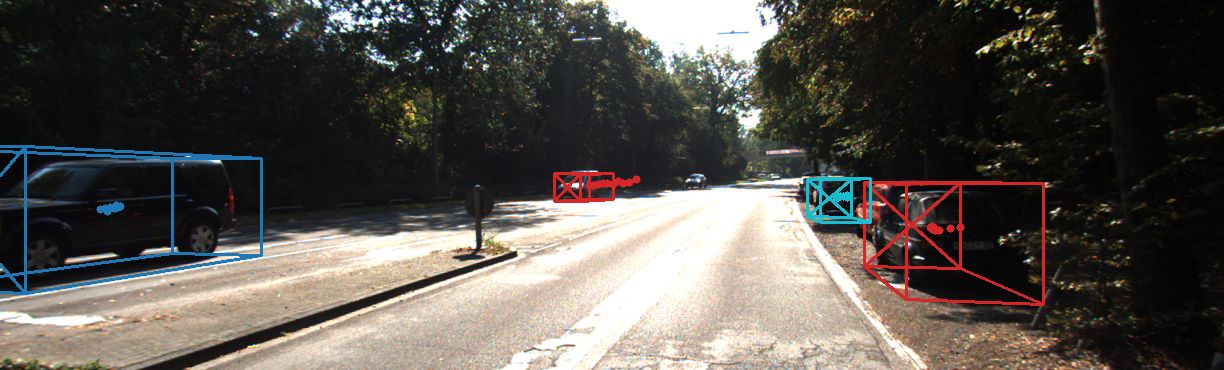} &  \includegraphics[width=\linewidth, height=1.5cm, frame]{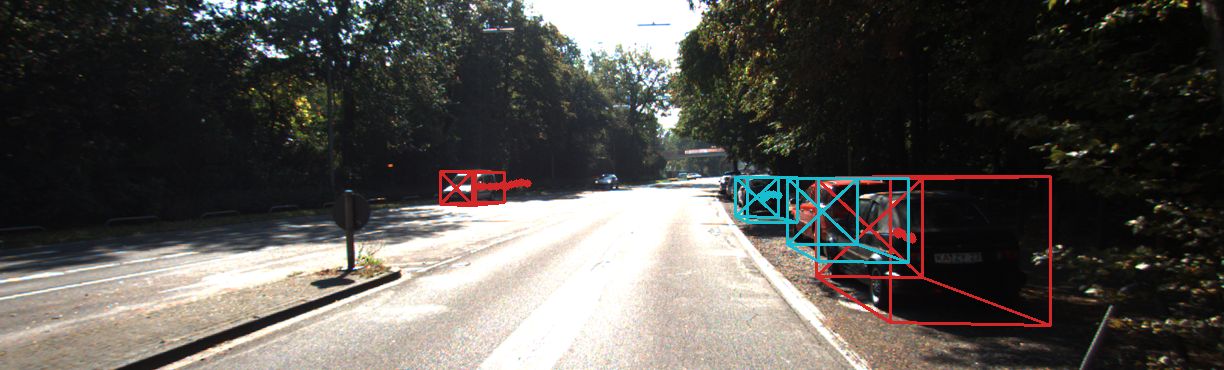} & \includegraphics[width=\linewidth, height=1.5cm, frame]{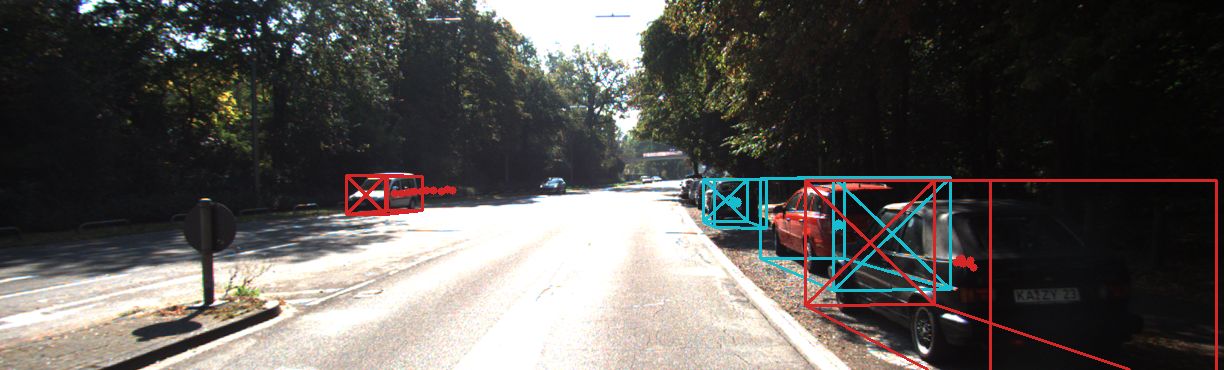}
\\
\\
\multirow{15}{*}{\rotatebox[origin=c]{90}{nuScenes}} & (d) & 
\includegraphics[width=\linewidth, height=1.5cm, frame]{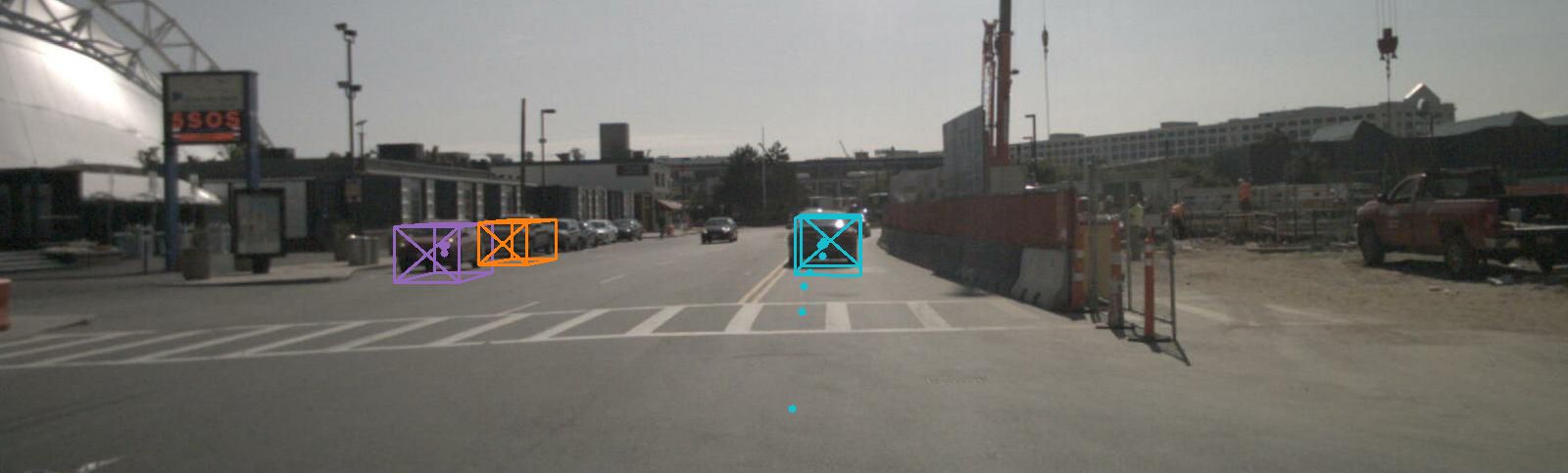} & \includegraphics[width=\linewidth, height=1.5cm, frame]{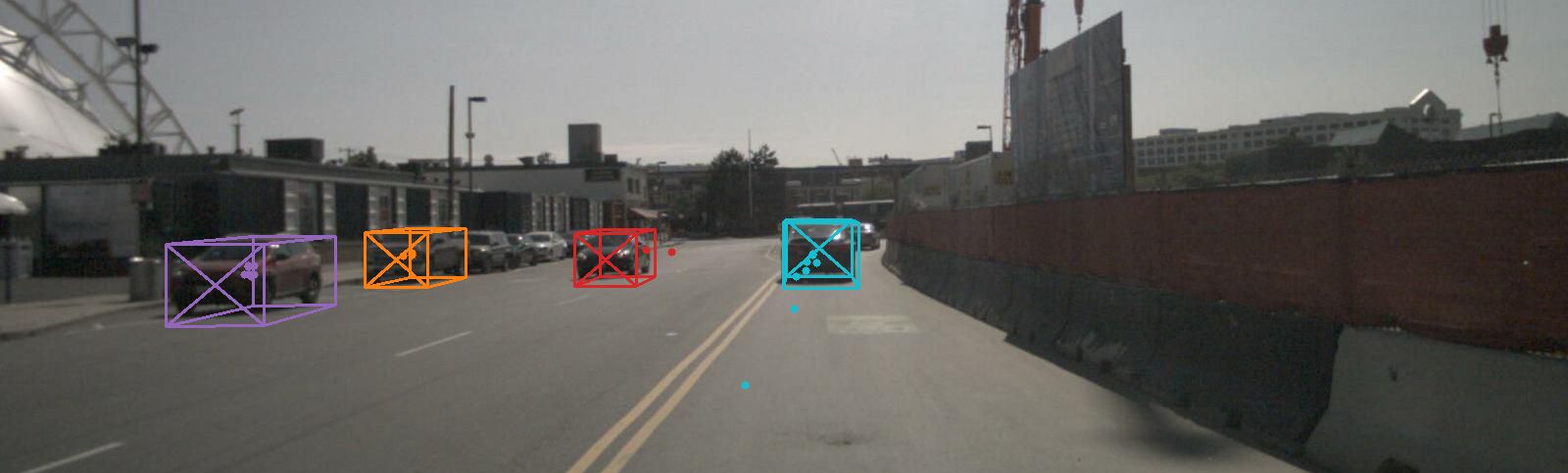} & \includegraphics[width=\linewidth, height=1.5cm, frame]{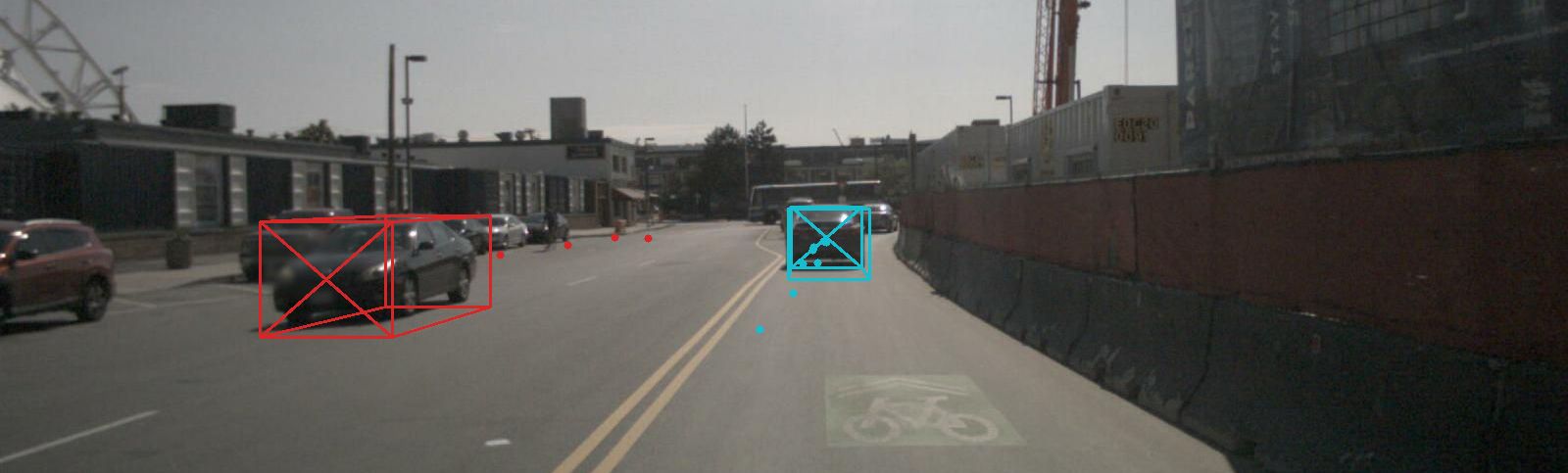}
\\
& (e) & 
\includegraphics[width=\linewidth, height=1.5cm, frame]{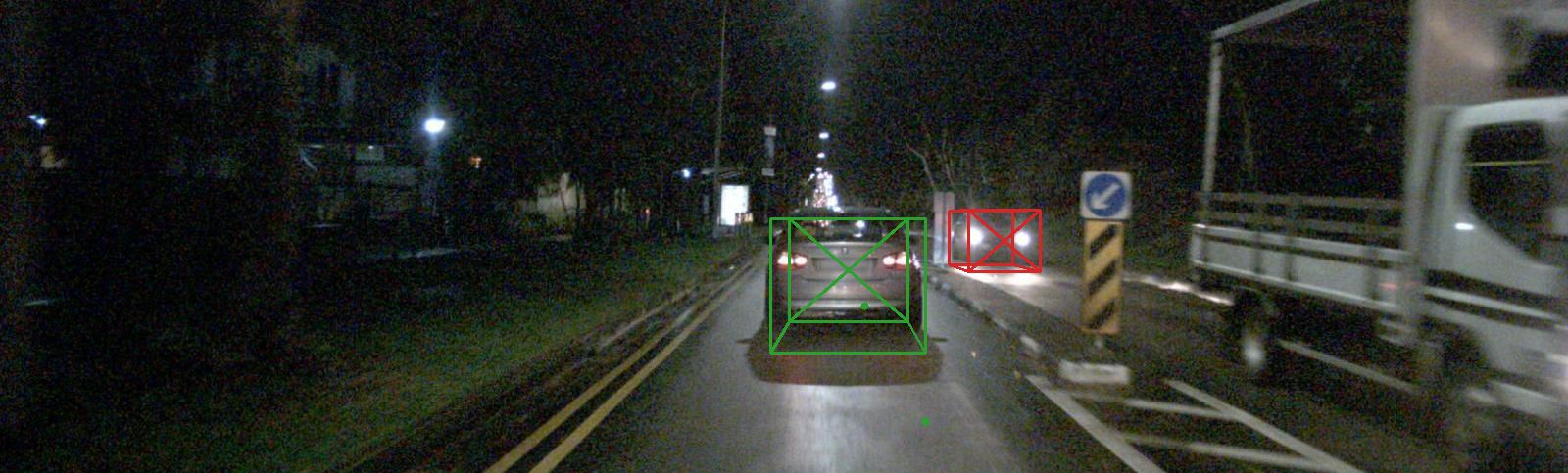} & \includegraphics[width=\linewidth, height=1.5cm, frame]{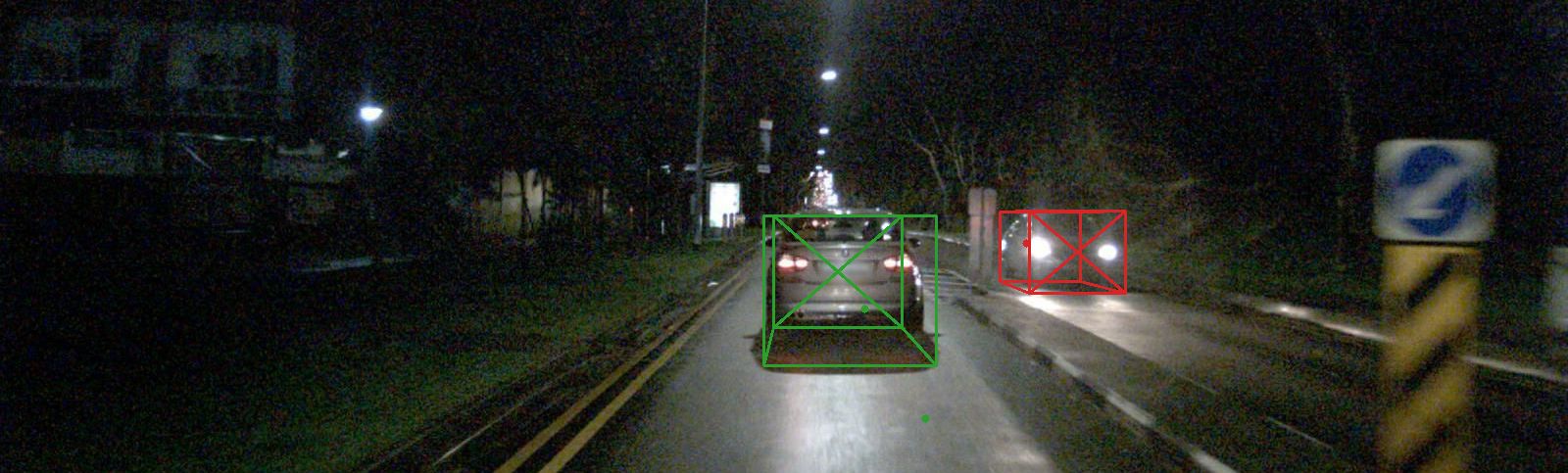} & \includegraphics[width=\linewidth, height=1.5cm, frame]{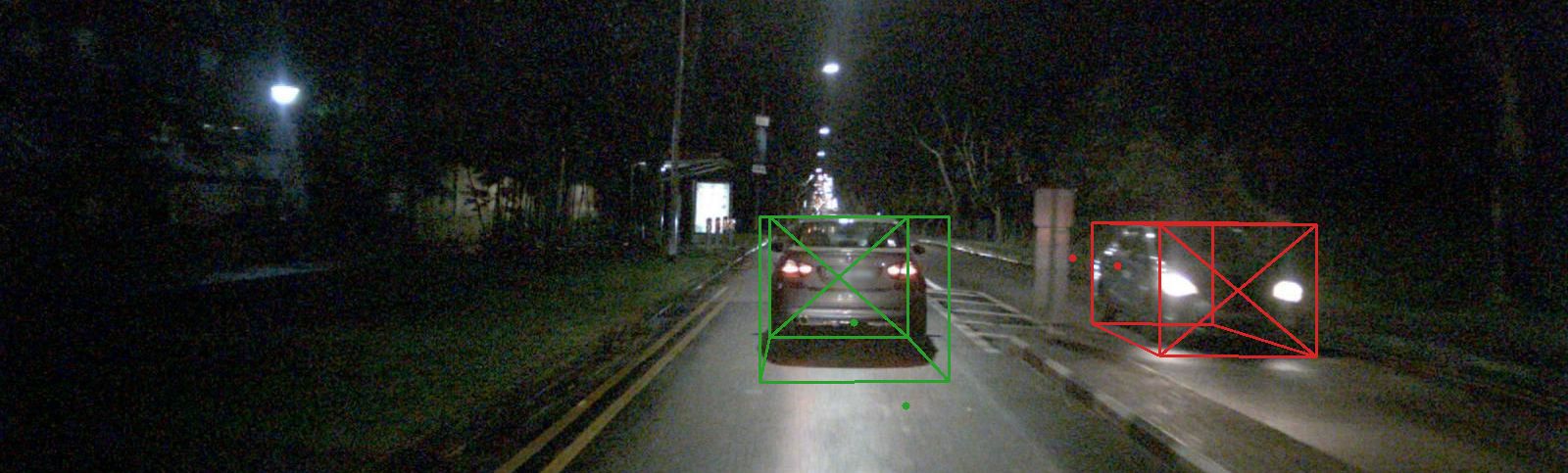}
\\
 & (f) & 
\includegraphics[width=\linewidth, height=1.5cm, frame]{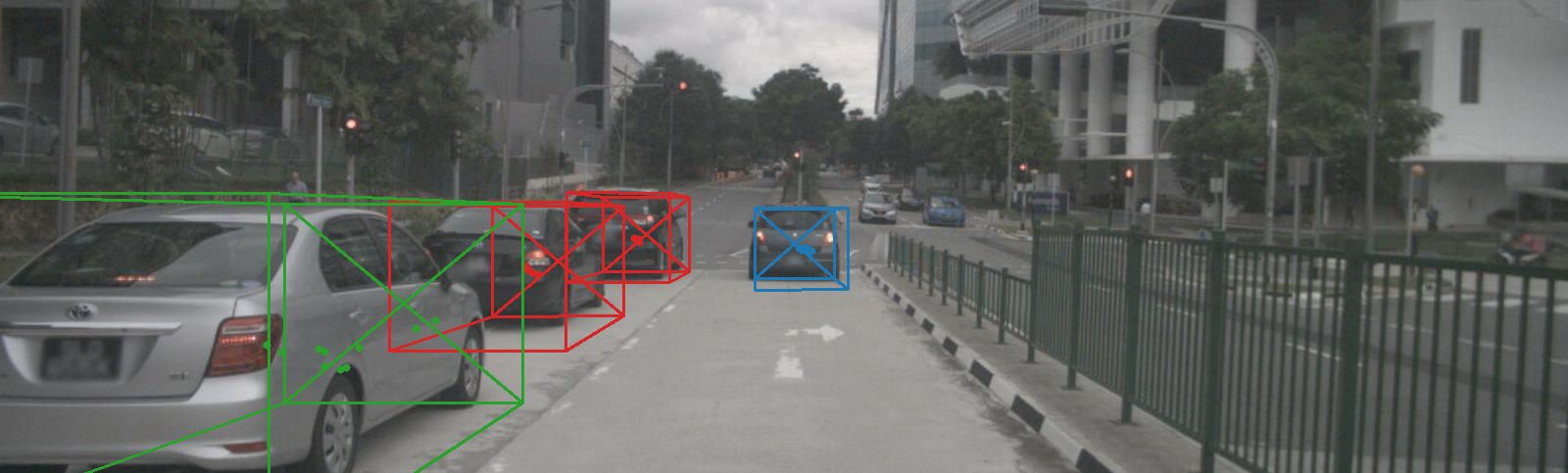} & \includegraphics[width=\linewidth, height=1.5cm, frame]{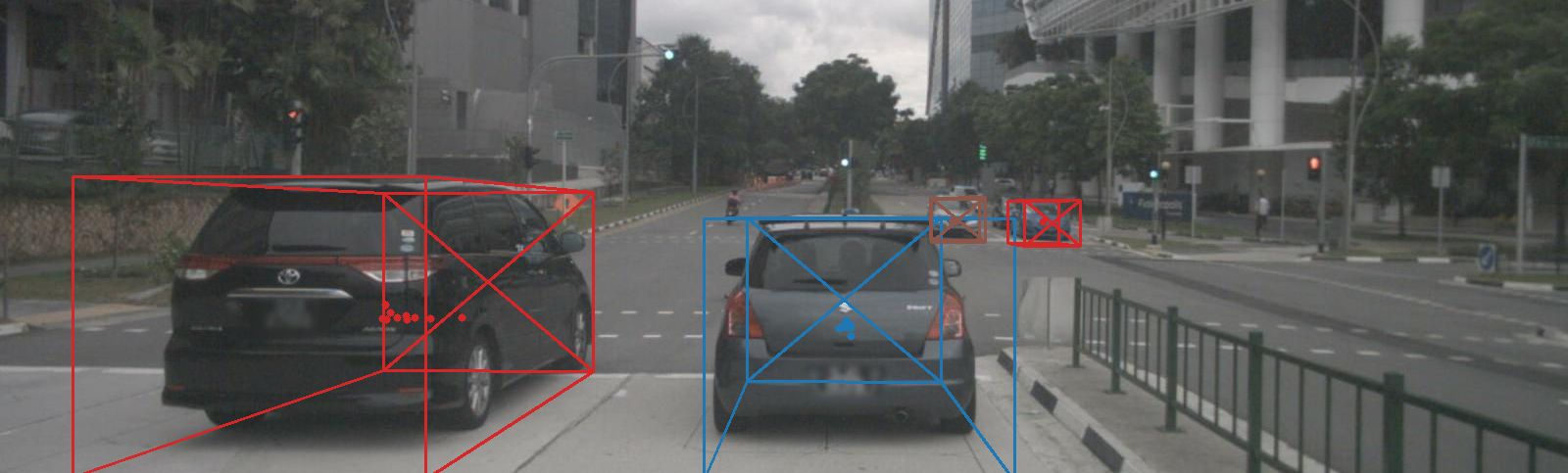} & \includegraphics[width=\linewidth, height=1.5cm, frame]{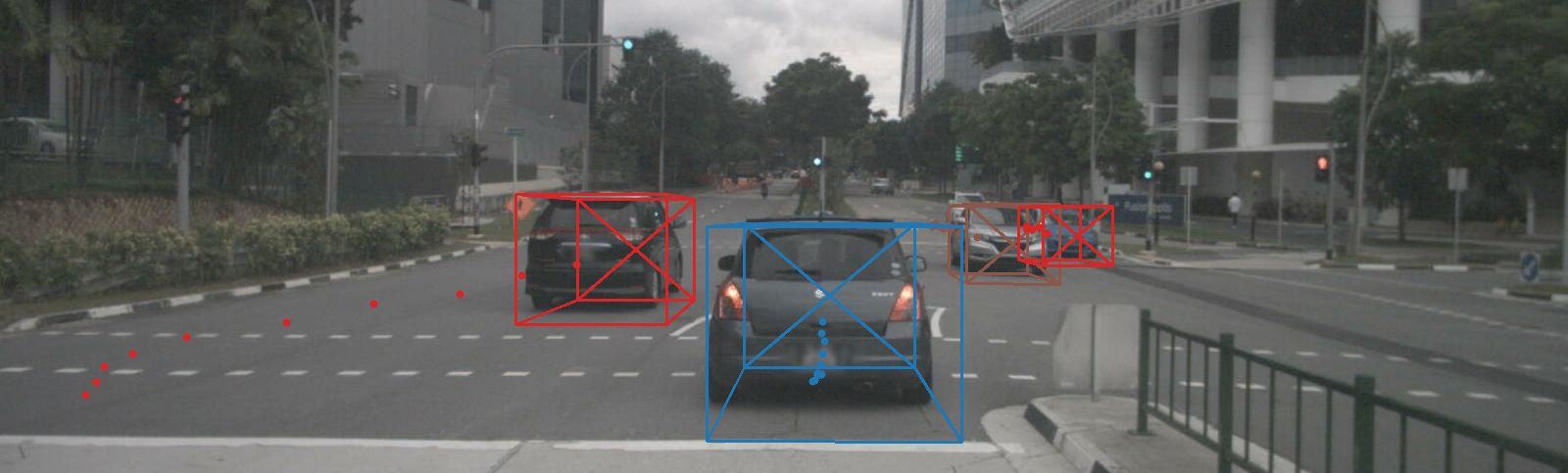}
\\
\end{tabular}
}
\caption{Visualization of the tracking results generated by the base network, CenterTrack, when trained using our 3D pseudolabels. Each object track is visualized in a unique color, and its previous trajectory is illustrated using a sequence of dots of the same color. Observe that CenterTrack generalizes to unseen image sequences and maintains accurate track predictions under occlusion, across varying geographic regions, and under different lighting conditions.}
\label{fig:supp-qual-centertrack}
\vspace{-0.3cm}
\end{figure}

\subsection{Base Network Predictions}
\label{subsec:qual-base-network}
In this section, we present additional qualitative results demonstrating the 3D tracking performance of the base network, CenterTrack, when trained with the 3D pseudolabels generated by \net. \figref{fig:supp-qual-centertrack} presents these results. 
We observe from \figref{fig:supp-qual-centertrack}(a, b, c) that CenterTrack exhibits strong generalization to previously unseen image sequences and produces high-quality 3D tracking predictions on the KITTI dataset. The network leverages our 3D pseudolabels to maintain track consistency across multiple objects in the scene, even when they are significantly occluded. The trajectory history of each object, visualized using a sequence of points, further highlights that CenterTrack preserves spatial coherence over extended time horizons. This indicates that the supervision signal derived from our pseudolabels effectively guides the learning of stable object dynamics. 

\figref{fig:supp-qual-centertrack}(d, e, f) demonstrate that the benefits of our 3D pseudolabel generation framework also extend to the more challenging nuScenes dataset. CenterTrack generates high-quality predictions across diverse geographic regions (USA and Singapore) and lighting conditions (day and night). These results thus demonstrate that \net~generates rich 3D pseudolabels from only sparse annotations, thereby allowing existing monocular 3D object trackers to be trained under extreme label sparsity.